\documentclass{article}
\usepackage[final]{colm2026_conference}

\usepackage{longtable}
\usepackage{array}
\usepackage{lscape}  
\usepackage{xcolor}
\usepackage{tcolorbox}
\usepackage{enumitem}
\tcbuselibrary{breakable,skins}
\usepackage{booktabs}
\usepackage{amsmath}
\usepackage{tikz}
\usepackage{float}   
\usepackage{hyperref}
\usepackage{url}
\usepackage{lineno}
\usepackage{booktabs}
\usepackage{pifont}
\usepackage{multirow}
\usepackage{tabularx}

\usepackage[T1]{fontenc}

\usepackage[utf8]{inputenc}
\usepackage{xspace}
\usepackage{microtype}

\usepackage{inconsolata}

\usepackage{graphicx}
 
\title{\textsc{ConvoLearn}: A Learning Sciences Grounded Dataset for Fine-Tuning Dialogic AI Tutors}

\author{Mayank Sharma \\
    Stanford University \\
    \texttt{masharma@stanford.edu} \\
    \And
    Roy Pea \\
    Stanford University \\
    \texttt{roypea@stanford.edu} \\
    \And
    Hari Subramonyam \\
    Stanford University \\
    \texttt{harihars@stanford.edu}
}

\begin{document}
\ifcolmsubmission
\linenumbers
\fi
\maketitle

\begin{abstract}

Despite their growing adoption in education, LLMs remain misaligned with the core principle of effective tutoring: the dialogic construction of knowledge. We introduce 
\textsc{ConvoLearn}\footnote{\url{https://huggingface.co/datasets/masharma/convolearn}}, a dataset of 2,134 semi-synthetic tutor-student 
dialogues operationalizing six dimensions of dialogic tutoring grounded 
in knowledge-building theory, situated in middle school 
Earth Science curriculum. We show that dimension-labeled dialogic 
training data captures meaningful pedagogical signal that generalizes 
beyond its semi-synthetic domain: scores from a classifier trained on 
\textsc{ConvoLearn} correlate significantly with expert-coded 
instructional quality in authentic classrooms across multiple subscales 
(range $r = .118$-.258, all $p < .05$). As a proof of concept, we fine-tune \textsc{Mistral-7B} on \textsc{ConvoLearn} and show that dimension-level fine-tuning can steer a 7B open-weight model toward dialogic tutoring behavior that credentialed teachers rate as competitive with a strong proprietary baseline. With this work, we support the development of AI tutors capable of more dialogic interactions.

\end{abstract}

\section{Introduction}

Large language models (LLMs) are rapidly reshaping the educational landscape. 
According to a nationally representative survey, 54\% of K-12 students in the 
United States reported using AI for school work during the 2024-25 school year 
\citep{doss2025ai}, and AI investment in education is projected to reach \$112 
billion by 2034 \citep{precedence2025ai}. Much of this adoption centers around AI-powered tutoring, where LLMs are being deployed 
to provide students with personalized academic support (for example, Khanmigo \citep{khanmigo}). This focus builds on longstanding evidence that individualized instruction can dramatically improve learning outcomes: \citet{bloom1984sigma}’s two-sigma study showed that 1:1 tutoring enabled average students to reach the 98th percentile of traditionally instructed peers. LLMs hold promise to approximate this personalized tutoring at scale.

Despite this optimism, research in educational NLP demonstrates that LLMs exhibit 
fundamental pedagogical limitations. Modern foundation models are alignment-tuned 
to be broadly `helpful' assistants, but helpfulness is often considered `at odds 
with pedagogy and learning' \citep{jurenka2024towards}. RLHF/RLAIF techniques 
further optimize for direct and satisfying answers \citep{ouyang2022training}, 
producing models that talk \textit{at} students rather than \textit{with} them. 
Effective tutoring, however, is inherently dialogic: it draws out student thinking, 
builds on it, and guides students to construct understanding themselves 
\citep{https://doi.org/10.1207/s15516709cog2504_1}. Research confirms this gap is large and consistent: despite scoring high on surface qualities like coherence and human-likeness \citep{maurya2025unifying}, LLMs substantially underperform human tutors on dialogic dimensions such as providing meaningful guidance \citep{maurya2025unifying} and engaging students in genuine reasoning and learning rather than passive receipt of information \citep{macina2023opportunities, tack2022ai}. 

The research community has responded with approaches such as Socratic prompting \citep{zhang2024splsocraticplaygroundlearning, liu2024socraticLM, favero2024enhancingcriticalthinkingeducation}, evaluation benchmarks \citep{macina2023opportunities, maurya2025unifying, kochmar-etal-2025-findings}, and most recently, fine-tuning datasets built from authentic tutoring recordings \citep{perczel2025teachlmposttrainingllmseducation}. Each addresses a different layer of the problem: prompting steers behavior at inference time without changing what a model has learned, benchmarks quantify pedagogical failures without providing a signal to correct them, and authentic recordings capture naturalistic tutoring behavior but are not decomposed into theoretically grounded dimensions that could guide targeted fine-tuning. While proprietary systems such as LearnLM attempt to directly address these limitations through pedagogy-informed training \citep{jurenka2024towards}, their data and processes are not publicly available. The result is a critical gap: no openly available dataset labels specific, theoretically grounded dialogic behaviors at the dimension level, leaving researchers without the training signal needed to steer models toward effective tutoring. This gap is all the more consequential given evidence that LLM misalignment with effective pedagogy is structural, rooted in shared pretraining rather than model or prompt choice \citep{hardy2026knowledgewisdommeasuringmisalignment}, making targeted, theory-driven fine-tuning data not just useful, but necessary.

We introduce \textsc{\textbf{ConvoLearn}}, a dataset of 2,134 dialogic tutor–student dialogues (\textasciitilde20 turns each), in which teacher turns are authored by credentialed human teachers and student turns are simulated. While \textsc{ConvoLearn} cannot fully overcome structural pretraining biases, it provides an openly available resource operationalizing six dimensions of dialogic tutoring grounded in knowledge-building theory \citep{Scardamalia_Bereiter_2014}: cognitive engagement, formative assessment, accountability, cultural responsiveness, metacognition, and power dynamics, across 21 fine-grained subdimensions, situated in California's middle school Earth Science curriculum. Our contributions are threefold: \textbf{(1) Dataset}: we release \textsc{ConvoLearn} with quality ratings preserved across the spectrum, enabling diverse downstream uses including fine-tuning, classifier training, and contrastive learning; \textbf{(2) Ecological validity}: we provide partial external validation for three dimensions most amenable to linguistic operationalization (cognitive engagement, metacognition, and formative assessment) by training a Longformer classifier on \textsc{ConvoLearn} and applying it to authentic classroom transcripts from the NCTE corpus \citep{demszky-hill-2023-ncte}, where derived scores correlate significantly with expert-coded instructional quality measures; and \textbf{(3) Proof of concept}: we fine-tune open-weight LLMs on a high-quality subset using multi-turn QLoRA and evaluate against proprietary baselines via teacher raters, demonstrating that dimension-level fine-tuning on \textsc{ConvoLearn} can steer models toward more dialogic tutoring strategies.

\section{Prior Work}

Much of the literature on AI tutors has centered on \textit{mistake remediation}, focusing on identifying and correcting student errors in structured problem-solving settings. Here, we instead focus on work related to \textit{dialogic tutoring}, an area that remains relatively underexplored and emphasizes eliciting student thinking, building on their ideas, and guiding learners to construct understanding themselves \citep{https://doi.org/10.1207/s15516709cog2504_1}. Early work by \citet{tack2022ai} assessed models on their ability to produce teacher talk, understand students, and help them, finding that conversational agents fell significantly short of human teachers on all three dimensions, even when performing comparably on surface measures such as conversational uptake. \citet{macina2023mathdialdialoguetutoringdataset} showed that, although models could produce fluent tutoring responses, human evaluation revealed substantially fewer genuine learning opportunities through scaffolding, reflection, and engagement. Building on these findings, \citet{maurya2025unifying} consolidated prior efforts into a unified eight-dimension taxonomy, demonstrating that even SoTA LLMs function more like Q\&A systems, scoring highly on qualities such as human-likeness while underperforming on guidance and actionability.

\subsection{Approaches to Dialogic Alignment}

Attempts to align LLMs toward more dialogic tutoring behaviors have included 
prompting, evaluation benchmarking, and fine-tuning. Socratic prompting 
approaches show that steering models toward inquiry-based dialogue can improve 
critical thinking and self-reflection 
\citep{zhang2024splsocraticplaygroundlearning, liu2024socraticLM, 
favero2024enhancingcriticalthinkingeducation}, though these modify behavior 
only at inference time without changing what a model has learned. On the 
evaluation side, \citet{maurya2025unifying} release MRBench, a benchmark of 
192 tutoring conversations annotated across eight pedagogical dimensions. 
Most recently, \citet{jiang2025sidbenchmarkingguidedinstruction} introduce 
SID, a benchmark of 10,000 interdisciplinary STEM dialogue turns specifically 
targeting dialogic guidance, finding that even state-of-the-art LLMs struggle 
to execute effective guided dialogues that support knowledge integration and 
transfer, establishing the evaluation gap but leaving the training data gap 
unaddressed. On the fine-tuning side, Google's LearnLM applies pedagogy-informed post-training to improve instructional behaviors 
such as guiding students to answers and promoting engagement 
\citep{jurenka2024towards}, but its data and training process are not publicly 
available. TeachLM, trained on over 100,000 hours of authentic one-on-one 
tutoring interactions using PEFT, increases student talk time, diversifies 
questioning strategies, and produces around 50\% more dialogue turns than 
base models \citep{perczel2025teachlmposttrainingllmseducation}, though its 
training data is likewise proprietary.

\subsection{Remaining Gaps}

Despite these advances, significant gaps remain, as summarized in 
Table~\ref{tab:dataset_comparison}. No existing fine-tuning resource is 
annotated at the level of specific, theoretically grounded dialogic dimensions: 
both LearnLM and TeachLM demonstrate the value of pedagogy-informed fine-tuning 
but make neither their training data nor dimension-level labels publicly 
available. While SID provides the closest analog to a dialogic tutoring 
benchmark, it is evaluation-only, fully synthetic, and grounded in a 
non-English curriculum, establishing the evaluation gap but leaving the 
training data gap unaddressed. Furthermore, no existing resource includes conversation-level quality ratings, limiting their utility for 
effectiveness-weighted fine-tuning and contrastive learning. 
\textsc{ConvoLearn} addresses these gaps by providing 2,134 dialogues labeled 
across six theoretically grounded dimensions of dialogic tutoring, authored by 
323 credentialed K-12 teachers with a mean of 10.9 yrs of experience, with 
full quality ratings preserved across the quality spectrum to support diverse 
downstream uses.

\begin{table}[h]
\centering
\small
\setlength{\tabcolsep}{4pt}
\begin{tabular}{lccccc}
\toprule
\textbf{Property} & \textbf{MRBench}$^*$ & \textbf{SID} & \textbf{LearnLM} & \textbf{TeachLM} & \textbf{\textsc{ConvoLearn} (Ours)} \\
\midrule
Size & 192 convos & 10K turns & N/A & 100K hrs & \textbf{2,134 convos} \\
Avg. Turns & -- & 5+ & -- & -- & \textbf{20} \\
Setting & K--12 & K-12 & Mixed & Mixed & \textbf{K--12} \\
Subject & Math & STEM & Mixed & Mixed & \textbf{Earth Sci.} \\
Type & Eval & Eval & FT & FT & \textbf{FT} \\
Dim. Labels & \ding{51} (8) & \ding{51} (9) & \ding{55} & \ding{55} & \textbf{\ding{51} (6+21)} \\
Quality Ratings & \ding{55} & \ding{55} & \ding{55} & \ding{55} & \textbf{\ding{51}} \\
Open & \ding{51} & \ding{51} & \ding{55} & \ding{55} & \textbf{\ding{51}} \\
\bottomrule
\end{tabular}
\caption{Comparison of \textsc{ConvoLearn} with related dialogic AI 
tutoring evaluation and fine-tuning resources. FT = Fine-tuning. 
Dim.\ Labels indicates whether dialogues are annotated with grounded 
pedagogical dimension labels (brackets indicate dimension count). 
Quality Ratings indicates whether per-dialogue effectiveness scores 
are included. Open indicates whether the resource is publicly 
available. $^*$MRBench is primarily focused on mistake remediation 
but includes several dimensions that overlap with dialogic tutoring 
concerns.}
\label{tab:dataset_comparison}
\end{table}

\section{Dialogic Learning Framework}

Effective tutoring is not merely about conveying correct information: it requires 
supporting learners in actively generating, refining, and connecting ideas. We 
ground \textsc{ConvoLearn} in knowledge-building theory 
\citep{Scardamalia_Bereiter_2014}, which frames learning as a collaborative 
process of constructing and refining understanding through discourse rather than 
passive receipt of information. This positions the tutor not as an 
answer-provider but as a dialogic partner who elicits student thinking and 
adapts to it in real time. For example, when a student asks ``\textit{Why did 
the dinosaurs disappear?}'', a knowledge-passing tutor simply states that an 
asteroid caused the extinction. A knowledge-building tutor instead opens space 
for inquiry: ``\textit{What ideas do you have about why dinosaurs may have 
disappeared?}'' This invites students to propose possibilities such as climatic 
shifts, food-web collapse, or volcanic activity, followed by dialogic moves that 
help them compare theories, evaluate evidence, and refine their explanations. We 
operationalize this framework through six dimensions of dialogic tutoring, further broken down in 21 subdimensions (see full definitions 
of each subdimension and dataset excerpts in Appendix~\ref{app:kb_dimensions}).

\textbf{Cognitive Engagement} refers to the depth of processing and quality 
of thinking strategies students employ during learning 
\citep{blumenfeld2006motivation, chi2014icap}. In dialogic tutoring, cognitive 
engagement is the most direct behavioral counterpart to answer-giving: where an 
answer-giving tutor resolves cognitive challenge by providing solutions, a 
dialogically engaging tutor uses that challenge as the site of learning. 
Linguistically, it manifests as open-ended questioning, uptake of student ideas, 
and scaffolded elaboration rather than declarative explanation. It is 
operationalized through four subdimensions: scaffolding, critical thinking, 
generative questioning, and problem-based reasoning.

\textbf{Formative Assessment} refers to the ongoing, interactive monitoring of student understanding during instruction to regulate learning in real time \citep{cowie1999model, black2009developing}. Unlike summative evaluation, it is embedded within the dialogue: tutors attend to student contributions, interpret them against learning goals, and adapt their next move accordingly. Linguistically, it appears as comprehension checks, probing follow-up questions, and responses that build on or correct student ideas. It is operationalized through three subdimensions: continuous assessment, self-assessment, and synthesizing.

\textbf{Accountability} reflects expectations that discourse aligns with 
norms of evidence and reasoning \citep{michaels2008deliberative}. In dialogic 
tutoring, accountability moves the conversation beyond mere exchange of opinions 
toward epistemic responsibility: students are expected to justify claims, 
evaluate evidence, and engage with counterarguments. Linguistically, it manifests 
as tutor prompts that require students to cite evidence, explain their reasoning, 
or defend a position. It is operationalized through three subdimensions: 
evidence-based reasoning, moral responsibility, and depth of reasoning.

\textbf{Cultural Responsiveness} recognizes that effective instruction must 
engage learners' diverse cultural backgrounds and build on their prior knowledge 
and experiences \citep{ladsonbillings1995toward, au1981teaching}. In dialogic 
tutoring, it requires tutors to connect concepts to 
culturally relevant contexts and affirm diverse ways of knowing rather than 
assuming a single canonical frame. Linguistically, it manifests as analogies 
drawn from students' cultural contexts and invitations to connect content 
to their experience. It is operationalized through two 
subdimensions: cultural identity exploration and cultural analogies.

\textbf{Metacognition} refers to awareness and regulation of one’s own cognitive processes \citep{flavell1981monitoring}. In dialogic tutoring, it involves prompting reflection and modeling through thinking aloud, making expert reasoning visible and encouraging students to adopt similar habits. Linguistically, it appears as prompts asking students to explain their reasoning, identify difficulties, or compare current understanding with prior beliefs. It is operationalized through six subdimensions: reflective growth, strategic thinking, self-reflection, thinking aloud, error analysis, and reflection.

\textbf{Power Dynamics} captures how agency and participation are distributed 
in learning interactions \citep{gordon1980foucault}. In dialogic tutoring, 
equitable power dynamics are essential: tutors who dominate the conversational 
floor or position themselves as sole arbiters of knowledge undermine the 
collaborative construction of understanding. Linguistically, it manifests in 
turn-taking patterns, the degree to which student ideas are taken up and built 
upon, and whether students are positioned as active contributors. It is operationalized through three subdimensions: ownership of ideas, 
partisanship, and persuasive discourse.

\section{Data Collection}

\begin{figure*}[t]
\centering
\includegraphics[width=\textwidth]{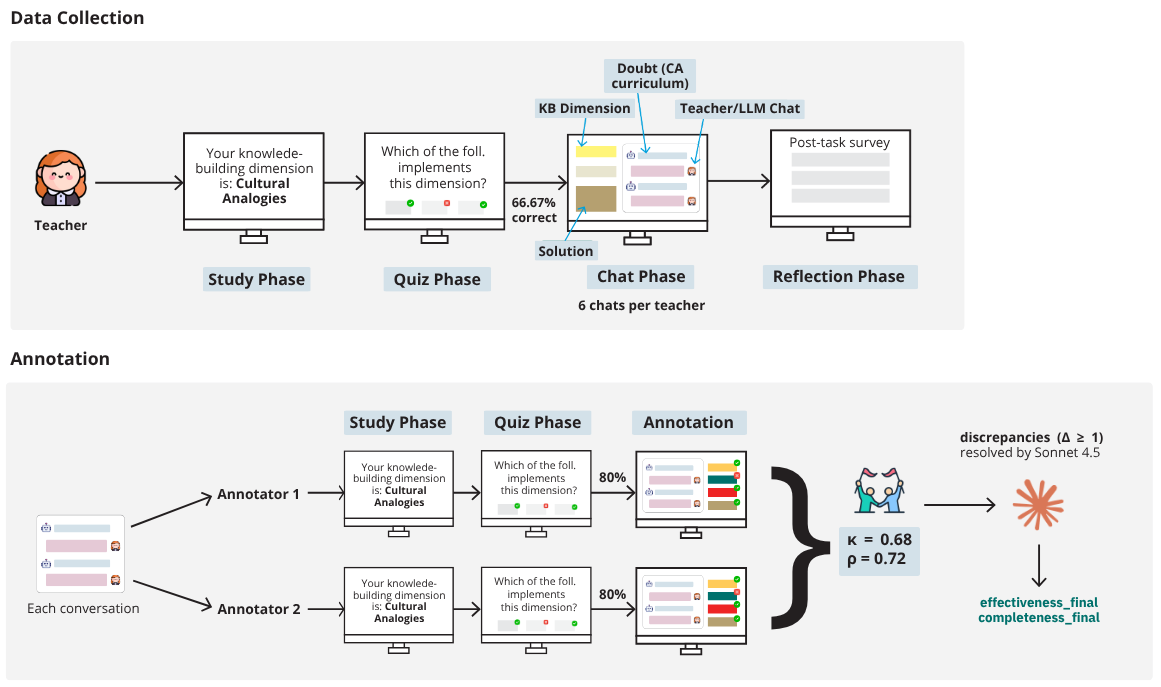}
\caption{Multi-step pipeline for data collection and annotation. Teachers interact with a simulated student on Earth Science prompts, producing dialogues that are independently scored by annotators, with discrepancies adjudicated by 
\textsc{Claude Sonnet~4.5}.}
\label{fig:wide_figure}
\end{figure*}

The data collection, annotation and filtering pipeline is represented pictorially in Figure~\ref{fig:wide_figure}.

\subsection{Question Selection}

All tutoring conversation starter prompts were derived from 60 multiple-choice questions from a freely available California Earth Science Standards Test for middle school students, spanning four domains: Investigation and Experimentation, Astronomy and Cosmology, Solid Earth, and Earth’s Energy Systems. Each question was rephrased as a first-person student “doubt” to initiate dialogue; for example, a question on continental drift became “What’s the best evidence that Earth’s continents were once positioned differently than they are today?”, ensuring each conversation began from a curriculum-aligned knowledge gap. Middle school Earth Science was chosen deliberately: it is conceptually rich and underserved by existing tutoring datasets, which tend to focus on mathematics and coding.

\subsection{Tutoring Interface and Student Simulation}

Prior to formal data collection, we conducted pilot sessions with 
two teachers to explore feasibility of 
implementing the six target dimensions. Participants completed 
fill-in-the-blank templates using student ``doubts'' in a think-aloud 
format, helping refine dimension definitions and test ease of tutoring on the dimensions. For the actual data collection, a custom web platform facilitated controlled tutor-student interactions (see screenshots in Appendix~\ref{app:tut_interface}). All 
tutors interacted with a simulated student (``Jamie'') powered by the 
\textsc{Gemini-1.5-Pro} API, chosen for its multi-turn reasoning and 
long-context reliability. A fixed system prompt enforced consistent student behavior (Appendix~\ref{app:student_sim_prompt}). Three internal pilots refined the interface, prompt design, and logging mechanisms. To assess sensitivity to the student simulator, we later evaluated the 
fine-tuned model using \textsc{GPT-4o} as an alternative simulator 
across three pedagogically distinct student profiles (engaged, limited 
prior knowledge, and disengaged), finding strong performance across all 
three (\textit{M} $> 4.0$; see Appendix~\ref{app:sim_sensitivity}), 
providing some evidence for generalizability to various student personas.

\subsection{Teacher Recruitment and Data Collection}

We recruited 500 U.S.-based certified teachers via Prolific. After 
completing consent procedures and a short demographics form, each teacher 
was randomly assigned two of 21 subdimensions, reviewed definitions 
and two example dialogues per subdimension, and completed a qualification 
quiz (minimum 2/3 per subdimension required to proceed). Passing teachers 
engaged in six tutoring conversations (three per subdimension, 20 turns 
each: 10 teacher, 10 student). Of 500 participants, 72 failed the quiz; 
323 completed all tasks (mean experience: 10.9 years; 60.7\% rated 
Jamie's realism 4 or 5 out of 5; full demographics in 
Appendix~\ref{app:data_collection_demo}). Non-passing teachers also
received some compensation for their time.

\subsection{Data Integrity}

A total of 3,076 conversations were extracted from the backend, as some 
recruited teachers dropped off mid-task. Multi-stage filtering was then 
employed: first, incomplete dialogues ($<$10 teacher turns) were removed, 
yielding 2,786 dialogues (90.6\%). The platform logged keystroke activity, 
computing average keystrokes per character to flag AI-generated responses; 
a threshold of $>$0.5 reduced conversations to 2,600 (threshold selection 
ablated in Appendix~\ref{app:keystrokes}). Removing duplicates and 
responses with errors yielded 2,155 final conversations. The final dataset spans all six dimensions with natural variation in 
coverage: metacognition (27.6\%), cognitive engagement (23.5\%), 
formative assessment (13.6\%), power dynamics (13.4\%), accountability 
(12.9\%), and cultural responsiveness (9.0\%).

\section{Dataset Annotation}

We employed a hybrid LLM annotation pipeline to label the 2,155 
collected dialogues at scale. Each dialogue was independently evaluated 
by \textsc{GPT-4o} and \textsc{Claude Haiku} after the same study and quiz phase as data collection using an identical 
structured rubric assessing: (i) \textit{effectiveness} (ii) 
\textit{completeness} (iii) 
\textit{quality issues} and (iv) 
\textit{safety issues} (see full protocol in 
Appendix~\ref{app:ann_protocol}). 

\noindent\textbf{Stage 1: Dual annotation.} Agreement between the two annotators was moderate (quadratic $\kappa = 0.68$, Spearman $\rho = 0.72$), consistent with the subjectivity of pedagogical judgment. \newline
\noindent\textbf{Stage 2: Safety filtering.} Dialogues flagged by either annotator for harmful content were removed, eliminating 21 conversations and yielding 2,134 safety-verified examples. \newline
\noindent\textbf{Stage 3: Quality filtering.} Dialogues were removed when both annotators independently flagged the same quality issue, excluding 616 conversations across categories including vagueness, confusion, repetition, technical errors, and artificiality, yielding 1,518 dialogues. \newline
\noindent\textbf{Stage 4: Consensus resolution.} When the two annotator 
ratings (effectiveness and completeness) differed by at most one point, 
we assigned the mean. Larger discrepancies were adjudicated by 
\textsc{Claude Sonnet~4.5}, which reviewed both annotator rationales 
and issued a final label, invoked in 10.2\% of effectiveness and 
2.9\% of completeness cases. While this introduces a potential 
circularity given Haiku's role as a primary annotator, the large majority of labels were assigned via 
mean-aggregation and never reached adjudication. We also acknowledge that LLM-based 
annotation introduces potential limitations and treat the resulting labels as 
silver-standard annotations. The high-quality subset used for 
fine-tuning was subsequently validated through independent human teacher evaluation.

\subsection{Dataset Statistics}

\textsc{ConvoLearn} is released in two configurations: the \textit{full 
dataset} of 2,134 conversations spanning the complete quality spectrum, 
supporting downstream uses including contrastive learning and 
effectiveness-weighted fine-tuning; and the \textit{high-quality subset} 
of 1,250 conversations, obtained by removing dialogues scoring below 
Effectiveness $<$ 3, Completeness $<$ 2, and quality issues, yielding a subset with mean Effectiveness 4.10/5 and 82.8\% 
fully complete conversations, used for fine-tuning experiments in this 
paper. Table~\ref{tab:dataset_stats} summarizes dimension coverage and 
quality statistics across both configurations. Although we targeted equitable collection across all dimensions, final 
distributions reflect variation in implementation difficulty: cultural 
responsiveness was hardest to 
operationalize authentically, as credentialed teachers struggled to 
connect content to a simulated student of unknown background. We retain 
it given its established importance to equitable tutoring.

\begin{table}[h]
\centering
\small
\resizebox{\columnwidth}{!}{%
\setlength{\tabcolsep}{4pt} 
\begin{tabular}{lcccccc|cc}
\toprule
& \multicolumn{6}{c}{\textbf{Dimensions}} & \multicolumn{2}{c}{\textbf{Overall}} \\
\cmidrule(lr){2-7} \cmidrule(lr){8-9}
\textbf{Dataset} & \textbf{Metacog.} & \textbf{Cog. Eng.} & \textbf{Form. Assess.} & \textbf{Power Dyn.} & \textbf{Account.} & \textbf{Cult. Resp.} & \textbf{Mean Eff.} & \textbf{\% Complete} \\
\midrule
\textbf{Full (2,134)} & 27.6\% & 23.5\% & 13.6\% & 13.4\% & 12.9\% & 9.0\% & 3.6 & 71.4\% \\
\textbf{HQ Subset (1,250)} & 24.5\% & 27.6\% & 15.6\% & 12.9\% & 13.1\% & 6.3\% & 4.10 & 82.8\% \\
\bottomrule
\end{tabular}%
}
\caption{Coverage and quality statistics for the full dataset and the HQ subset.}
\label{tab:dataset_stats}
\end{table}

\section{Ecological Validity}

It has become a growing concern in AI benchmarking and evaluation communities that benchmark performance fails to reflect real-world utility: models that score well on curated datasets often rely on surface patterns and fail to exhibit context-sensitive behaviors required in deployment \citep{eriksson2025trustaibenchmarksinterdisciplinary, 10.1145/3442188.3445922}. For 
educational datasets specifically, this risk is acute: a dataset 
grounded in semi-synthetic interactions could encode surface-level 
patterns of teacher talk without capturing the pedagogical substance 
that makes dialogic tutoring effective. To assess whether 
\textsc{ConvoLearn} captured meaningful pedagogical signal, we apply a Longformer classifier \citep{beltagy2020longformerlongdocumenttransformer} trained on \textsc{ConvoLearn} to authentic K-12 classroom transcripts 
from the NCTE corpus \citep{demszky-hill-2023-ncte}. NCTE is one of 
the few publicly available datasets that links naturalistic classroom 
discourse to expert-coded observation measures, making it a rare and valuable resource for external validation 
of our constructs \citep{demszky-hill-2023-ncte, hill2008mathematical}.

We focus on three dimensions: cognitive engagement, formative assessment, and metacognition, because they are most observable in teacher language. Cognitive engagement appears in question types and uptake, formative assessment in responsive feedback, and metacognition in reflective prompts, all recoverable from transcripts. The remaining dimensions (accountability, cultural responsiveness, and power dynamics) rely on contextual and relational cues that transcripts alone cannot capture, making linguistic classifiers unsuitable for validation at this stage. We examine correlations with two validated observation instruments: 
MQI \citep{hill2008mathematical}, which codes the cognitive demand 
and discourse quality of mathematics instruction, and CLASS 
\citep{pianta2008classroom}, which captures broader dialogic and 
relational teaching quality (see Appendix~\ref{app:ncte_subscales}). We acknowledge two structural mismatches: NCTE captures whole-classroom mathematics instruction while \textsc{ConvoLearn} targets one-on-one Earth Science tutoring. However, this mismatch serves as a rigorous test of generalizability. If scores derived from semi-synthetic Earth Science dialogues align with expert-coded ratings of authentic mathematics classrooms, it provides compelling evidence of a genuine, shared pedagogical signal that extends beyond within-domain validation.

\subsection{Training}

We fine-tuned a Longformer regression model on \textsc{ConvoLearn} to predict conversation-level effectiveness scores, using dimension-tagged dialogues and consensus scores as targets. Dialogues were partitioned by seed question (70/15/15 train/val/test) to prevent leakage, and the model was initialized from \texttt{allenai/longformer-base-4096} and trained for 5 epochs. On the held-out test set, it achieved Pearson $r = 0.736$ ($p < .001$), RMSE $= 0.710$, and MAE $= 0.530$. We then applied the model to NCTE transcripts by segmenting them into non-overlapping 20-turn chunks, scoring each chunk, and averaging to the teacher-year level before merging with MQI and CLASS scores ($n = 322$).

\subsection{Results}

Dimension scores show small but statistically significant positive 
associations with five of six MQI subscales and two of five CLASS 
subscales (Table~\ref{tab:validity}). Among MQI subscales, the 
strongest associations are with ETCA (Enacted Task Cognitive 
Activation), which codes whether teachers elicit genuine student 
reasoning rather than procedural recall ($r_{\text{CE}} = .258$, 
$r_{\text{FA}} = .250$, $r_{\text{MC}} = .224$, all $p < .001$); 
SMQR (Student Mathematical Questioning and Reasoning), which codes 
the degree to which students engage in mathematical argumentation 
($r_{\text{CE}} = .243$, $r_{\text{FA}} = .231$, $r_{\text{MC}} = 
.199$, all $p < .001$); and EXPL (Teacher Explanations), which captures how well teachers communicate 
mathematical reasoning ($r_{\text{CE}} = .219$, $r_{\text{FA}} = 
.213$, $r_{\text{MC}} = .194$, all $p < .001$). Smaller but 
significant associations are observed with MLANG (Mathematical 
Language) and LINK (Linking and Connections), both of which capture discourse quality beyond 
procedural correctness. MGEN (Developing Mathematical Generalization) is not significant across any dimension, consistent with its known 
label sparsity in the NCTE corpus. Among CLASS subscales, CLAPS 
(Applied Problem Solving) and CLQF (Quality of Feedback) reach 
significance for cognitive engagement and formative assessment but 
not metacognition ($r_{\text{CE}} = .130$, $r_{\text{FA}} = .125$, 
both $p < .05$), consistent with metacognition being harder to 
detect in whole-classroom discourse where reflective prompts are 
less frequent than in one-on-one tutoring. The three remaining 
CLASS subscales: CLCU (Content Understanding), CLSTENG (Student 
Engagement), and CLINSTD (Instructional Dialogue),  do not reach 
significance, likely reflecting that these capture broader instructional dimensions less directly tied to the turn-by-turn dialogic moves \textsc{ConvoLearn} operationalizes.

\begin{table*}[t]
\centering
\small
\begin{tabular}{llccccccc}
\toprule
& & \multicolumn{2}{c}{\textbf{FA}} & 
  \multicolumn{2}{c}{\textbf{CE}} & 
  \multicolumn{2}{c}{\textbf{MC}} \\
\cmidrule(lr){3-4} \cmidrule(lr){5-6} \cmidrule(lr){7-8}
\textbf{Instrument} & \textbf{Subscale} & $r$ & $p$ & $r$ & $p$ & $r$ & $p$ \\
\midrule
\multirow{6}{*}{MQI} 
  & ETCA   & \textbf{.250} & \textbf{$<$.001***} & \textbf{.258} & \textbf{$<$.001***} & \textbf{.224} & \textbf{$<$.001***} \\
  & SMQR   & \textbf{.231} & \textbf{$<$.001***} & \textbf{.243} & \textbf{$<$.001***} & \textbf{.199} & \textbf{$<$.001***} \\
  & EXPL   & \textbf{.213} & \textbf{$<$.001***} & \textbf{.219} & \textbf{$<$.001***} & \textbf{.194} & \textbf{$<$.001***} \\
  & MLANG  & \textbf{.161} & \textbf{.004**}     & \textbf{.171} & \textbf{.002**}     & \textbf{.140} & \textbf{.012*}     \\
  & LINK   & \textbf{.133} & \textbf{.017*}      & \textbf{.141} & \textbf{.012*}      & \textbf{.118} & \textbf{.035*}     \\
  & MGEN   & .070          & .210                & .079          & .157                & .048          & .394               \\
\midrule
\multirow{5}{*}{CLASS} 
  & CLAPS  & \textbf{.125} & \textbf{.024*}      & \textbf{.130} & \textbf{.019*}      & .107          & .055               \\
  & CLQF   & \textbf{.119} & \textbf{.033*}      & \textbf{.125} & \textbf{.025*}      & .096          & .087               \\
  & CLCU   & .084          & .133                & .093          & .097                & .063          & .261               \\
  & CLSTENG& .103          & .065                & .107          & .055                & .091          & .103               \\
  & CLINSTD& .031          & .574                & .039          & .487                & .008          & .890               \\
\bottomrule
\end{tabular}
\caption{Pearson correlations between \textsc{ConvoLearn} Longformer 
scores and NCTE instructional quality measures at the teacher-year 
level ($n = 322$). 
FA = Formative Assessment; CE = Cognitive Engagement; 
MC = Metacognition.
* $p < .05$, ** $p < .01$, *** $p < .001$.}
\label{tab:validity}
\end{table*}

\section{Fine-Tuning as Proof of Concept}

As a proof of concept, we fine-tuned three instruction-tuned open-source LLMs of similar 
size: \textsc{Qwen-2.5-7B-Instruct} 
\citep{qwen2025qwen25technicalreport}, 
\textsc{Llama-3.1-8B-Instruct} 
\citep{grattafiori2024llama3herdmodels}, and 
\textsc{Mistral-7B-Instruct-v0.3} \citep{jiang2023mistral7b} using 
QLoRA \citep{dettmers2023qloraefficientfinetuningquantized} on a 
single NVIDIA A100 GPU, trained on the high-quality subset with a 
dimension-stratified 85/15 train--test split (full training setup in 
Appendix~\ref{app:finetuning}). Each dialogue was converted into 
progressive training samples where each teacher turn $t_i$ is 
predicted given all preceding context $(t_0 \ldots t_{i-1})$. An 
auxiliary \textsc{RoBERTa} classifier trained on 2,134 
safety-verified dialogues was used to select the best-performing 
fine-tuned model (\textsc{Mistral-7B}) for human evaluation ("intrinsic" model 
selection details in Appendix~\ref{app:auxclass}). Note that this section is intended as a proof of concept rather than a comprehensive model comparison; the evaluation is small-scale with a limited rater sample.

\subsection{Teacher Evaluation}

Using 72 seed questions not seen during training (also from CA curriculum): 42 from Earth Science (7 per dimension) and 30 from Physics (5 per dimension), we generated four-turn dialogues between fine-tuned \textsc{Mistral-7B} and simulated student "Jamie" using the same pipeline as the intrinsic evaluation (Appendix~\ref{app:auxclass}). We then generated similar dialogues with \textsc{Claude 
Sonnet 4.5} and \textsc{Gemini 2.0 Flash}, each receiving only the 
assigned dimension name and task instruction (see Appendix ~\ref{app:extrinsic_prompt} for prompts and Appendix~\ref{app:qualitative} for sample dialogues generated by the 3 models). We recruited 31 independent certified teacher 
raters via Prolific who reviewed 4-5 randomly sampled dialogues per condition. 
Raters were shown the seed question, the dimension definition, and 
three anonymized dialogue responses (blinded and randomized); they rated each response on a 
1-5 effectiveness scale. 

A 
linear mixed-effects model with model type and dimension as 
fixed effects and random intercepts for participants was used for analysis. Results indicate that fine-tuned \textsc{Mistral-7B} performs comparably to \textsc{Claude 
Sonnet 4.5} overall ($M = 3.49$ vs. $M = 3.56$; $\beta = -0.07$, 
$SE = 0.12$, $z = -0.55$, $p = .583$), suggesting that dimension-level fine-tuning on a modest dataset can begin to close the dialogic alignment gap. \textsc{Gemini 2.0 
Flash} scores significantly higher overall ($M = 3.82$; $\beta = 
0.27$, $SE = 0.12$, $z = 2.25$, $p = .025$ vs. \textsc{Mistral}), 
consistent with its known teaching strengths following LearnLM 
integration \citep{jurenka2024towards}. Other detailed results are in Appendix~\ref{app:extrinsic_eval}.

\section{Implications and Future Work}

Three broad findings emerge from this work. First, we demonstrated that dialogic alignment 
of LLM tutors can be improved through targeted training data, which we positioned as 
a necessity beyond evaluation benchmarks that fail to provide corrective 
signals and prompting approaches that modify behavior at inference time 
without changing what models learn. Second, such data can capture 
meaningful pedagogical signal even when semi-synthetic: scores derived 
from one-on-one Earth Science tutoring correlate significantly with 
expert-coded instructional quality in authentic mathematics classrooms, 
providing evidence that theoretically-grounded training data can 
demonstrate ecological validity beyond its collection context. Third, 
acting on this signal is feasible at modest scale: fine-tuning on 1,250 dimension-labeled dialogues is sufficient to produce dialogic behavior that credentialed teachers rate as comparable to a strong proprietary baseline, offering some evidence that the pedagogoic gap documented is partially addressable through fine-tuning.

Our data collection and annotation pipeline demonstrates that 
dimension-labeled dialogic data can be collected at scale from 
credentialed teachers without prohibitive cost or expertise barriers: 
323 teachers with a mean of 10.9 years of experience produced 2,134 
dialogues across six theoretically-grounded dimensions, with 60.7\% 
rating the simulated student as realistic. That said, dimensions 
requiring contextual or relational cues (notably, cultural 
responsiveness and accountability, where teachers struggled to connect 
content to a simulated student of unknown background) proved harder 
to operationalize, and their underrepresentation in the final dataset 
is itself a finding about the limits of this process.

The ecological validity results warrant closer attention than their 
modest effect sizes might suggest. Correlations of $r = .118$-$.258$ 
are small in absolute terms but consistent across five of six MQI 
subscales and two CLASS subscales, and the cross-domain nature of the 
validation, Earth Science tutoring predicting mathematics classroom 
quality, makes surface confounds a less plausible explanation than a 
within-domain result would allow. The uniform null for MGEN, a subscale 
with known label sparsity in NCTE, further suggests the classifier 
responds to the underlying construct rather than incidental corpus 
features. Construct validity is nonetheless established for only three 
of six dimensions; accountability, cultural responsiveness, and power 
dynamics depend heavily on contextual, relational, and identity-based 
cues that transcript text cannot fully encode, and their validation will 
likely require authentic student populations, longitudinal data, or 
multimodal signals.

\textsc{Gemini~2.0~Flash}'s significant advantage over both fine-tuned 
\textsc{Mistral-7B} and \textsc{Claude Sonnet~4.5} is theoretically 
informative. Its edge is plausibly attributable to LearnLM's 
pedagogy-informed post-training at scale \citep{jurenka2024towards}, 
a fundamentally different regime from imitation fine-tuning on 1,250 
dialogues. This suggests dimension-labeled supervised data closes a 
substantial portion of the alignment gap, but the remaining distance 
may require preference-based feedback or larger data volume, both 
tractable extensions using \textsc{ConvoLearn}'s full quality spectrum 
for DPO or RLHF-style training.

Several limitations bound these conclusions. Construct validity is established for only three of six dimensions, leaving the other three theoretically rather than empirically grounded. Annotation relied on LLMs in the pipeline producing silver-standard labels. The dataset covers a single curricular domain and grade band, limiting generalizability across subjects, languages, and age groups. The dataset exhibits a severe class imbalance; dimensions such as Metacognition and Cognitive Engagement account for over half of the data, potentially limiting the model's ability to effectively learn underrepresented dimensions. Imitation fine-tuning teaches surface form rather than pedagogical purpose: the model learns to reproduce dialogic teacher talk without necessarily internalizing when or why a given move is appropriate. The PoC evaluation is small-scale and should not be interpreted as a comprehensive comparison of model capabilities. Finally, the link between dialogic behavior and actual student learning outcomes remains unestablished in this work.

\section*{Ethical Considerations}
All human evaluations adhered to institutional ethical standards with 
prior IRB approval. Teachers recruited via Prolific were compensated at 
or above minimum wage under transparent consent and debriefing 
procedures. All conversational examples were screened to remove 
sensitive, offensive, or potentially biased language, and no minors 
were involved in data collection. This study aims to improve the 
pedagogical capabilities of LLM-based tutoring systems, not to replace 
human teachers. We nonetheless recognize potential dual-use risks. 
Automated LLM tutors deployed at scale may exacerbate existing 
educational inequities if access is unevenly distributed, and 
imitation-based models that reproduce surface patterns of effective 
teaching without genuine pedagogical understanding risk providing 
students with interactions that appear dialogic but fail to support 
real learning. There is also a broader risk of over-automation: as 
AI tutors become more capable, institutional pressures to reduce human 
teaching roles may intensify in ways that undermine the relational, 
motivational, and equity-oriented dimensions of education that resist 
purely linguistic modeling. To mitigate these risks, all data and model 
outputs are intended solely for research on educational alignment, and 
we encourage future deployment work to evaluate impacts on learning 
outcomes and equity before scaling.

\bibliography{custom}

@techreport{doss2025ai,
  author = {Doss, C. J. and Bozick, R. and Schwartz, H. L. and Chu, L. and Rainey, L. R. and Woo, A. and Reich, J. and Dukes, J.},
  title = {AI use in schools is quickly increasing but guidance lags behind: Findings from the RAND survey panels},
  institution = {RAND Corporation},
  year = {2025},
  month = {September},
  number = {RR-A4180-1},
  url = {https://www.rand.org/pubs/research_reports/RRA4180-1.html}
}

@misc{precedence2025ai,
  author = {{Precedence Research}},
  title = {AI in Education Market Size to Surge USD 112.30 Bn by 2034},
  year = {2025},
  month = {July},
  url = {https://www.precedenceresearch.com/ai-in-education-market}
}

@article{bloom1984sigma,
  author = {Bloom, B. S.},
  title = {The 2 Sigma Problem: The Search for Methods of Group Instruction as Effective as One-to-One Tutoring},
  journal = {Educational Researcher},
  volume = {13},
  pages = {4--16},
  year = {1984},
  doi = {10.3102/0013189X013006004}
}

@incollection{blumenfeld2006motivation,
  author = {Blumenfeld, P. C. and Kempler, T. M. and Krajcik, J. S.},
  title = {Motivation and cognitive engagement in learning environments},
  booktitle = {The Cambridge handbook of the learning sciences},
  editor = {Sawyer, R. K.},
  publisher = {Cambridge University Press},
  pages = {475--488},
  year = {2006},
  doi = {10.1017/CBO9780511816833}
}

@article{chi2014icap,
  author = {Chi, M. T. H. and Wylie, R.},
  title = {The ICAP framework: Linking cognitive engagement to active learning outcomes},
  journal = {Educational Psychologist},
  volume = {49},
  number = {4},
  pages = {219--243},
  year = {2014},
  doi = {10.1080/00461520.2014.965823}
}

@article{ladsonbillings1995toward,
  author = {Ladson-Billings, G.},
  title = {Toward a theory of culturally relevant pedagogy},
  journal = {American Educational Research Journal},
  volume = {32},
  number = {3},
  pages = {465--491},
  year = {1995}
}

@incollection{au1981teaching,
  author = {Au, K. and Jordan, C. and others},
  title = {Teaching reading to Hawaiian children: Finding a culturally appropriate solution},
  booktitle = {Culture and the bilingual classroom: Studies in classroom ethnography},
  pages = {139--152},
  year = {1981}
}

@book{gordon1980foucault,
  author = {Gordon, C. and Foucault, M.},
  title = {Selected interviews and other writings 1972-1977},
  publisher = {Pantheon Books},
  address = {New York},
  year = {1980}
}

@incollection{flavell1981monitoring,
  author = {Flavell, J. H.},
  title = {Monitoring social cognitive enterprises: Something else that may develop in the area of social cognition},
  booktitle = {Social cognitive development: Frontiers and possible futures},
  volume = {11},
  pages = {272--287},
  year = {1981}
}

@article{michaels2008deliberative,
  author = {Michaels, S. and O'Connor, C. and Resnick, L. B.},
  title = {Deliberative discourse idealized and realized: Accountable talk in the classroom and in civic life},
  journal = {Studies in Philosophy and Education},
  volume = {27},
  pages = {283--297},
  year = {2008}
}

@misc{jurenka2024towards,
  title = {Towards responsible development of generative AI for education: An evaluation-driven approach},
  author = {Jurenka, I. and Kunesch, M. and McKee, K. R. and Gillick, D. and Zhu, S. and Wiltberger, S. and Phal, S. M. and Hermann, K. and Kasenberg, D. and Bhoopchand, A. and Gottweis, J. and Mikulik, V. and Fagan, F. and Novikov, A. and Kumar, A. and Piot, B. and Terzi, J. and Wang, C. and Elster, C. and others and Ramasesh, V. V.},
  year = {2024},
  eprint = {2407.12687},
  archivePrefix = {arXiv},
  url = {https://arxiv.org/abs/2407.12687}
}

@inproceedings{macina2023opportunities,
  author = {Macina, J. and Daheim, N. and Wang, L. and Sinha, T. and Kapur, M. and Gurevych, I. and Sachan, M.},
  title = {Opportunities and challenges in neural dialog tutoring},
  booktitle = {Proceedings of the 17th Conference of the European Chapter of the Association for Computational Linguistics},
  pages = {2357--2372},
  year = {2023},
  publisher = {Association for Computational Linguistics},
  doi = {10.18653/v1/2023.eacl-main.173}
}

@inproceedings{maurya2025unifying,
  author = {Maurya, K. K. and Srivatsa, K. A. and Petukhova, K. and Kochmar, E.},
  title = {Unifying AI tutor evaluation: An evaluation taxonomy for pedagogical ability assessment of LLM-powered AI tutors},
  booktitle = {Proceedings of the 2025 Conference of the Nations of the Americas Chapter of the Association for Computational Linguistics: Human Language Technologies (Volume 1: Long Papers)},
  pages = {1234--1251},
  year = {2025},
  publisher = {Association for Computational Linguistics},
  doi = {10.18653/v1/2025.naacl-long.57}
}

@inproceedings{ouyang2022training,
  author = {Ouyang, L. and Wu, J. and Jiang, X. and Almeida, D. and Wainwright, C. L. and Mishkin, P. and Zhang, C. and Agarwal, S. and Slama, K. and Ray, A. and Schulman, J. and Hilton, J. and Kelton, F. and Miller, L. and Simens, M. and Askell, A. and Welinder, P. and Christiano, P. and Leike, J. and Lowe, R.},
  title = {Training language models to follow instructions with human feedback},
  booktitle = {Advances in Neural Information Processing Systems},
  volume = {35},
  pages = {27730--27744},
  year = {2022},
  url = {https://arxiv.org/abs/2203.02155}
}

@inproceedings{tack2022ai,
  author = {Tack, A. and Piech, C.},
  title = {The AI teacher test: Measuring the pedagogical ability of Blender and GPT-3 in educational dialogues},
  booktitle = {Proceedings of the 15th International Conference on Educational Data Mining},
  pages = {522},
  year = {2022},
  publisher = {International Educational Data Mining Society}
}

@misc{perczel2025teachlmposttrainingllmseducation,
  title = {TeachLM: Post-Training LLMs for Education Using Authentic Learning Data},
  author = {Janos Perczel and Jin Chow and Dorottya Demszky},
  year = {2025},
  eprint = {2510.05087},
  archivePrefix = {arXiv},
  primaryClass = {cs.CL},
  url = {https://arxiv.org/abs/2510.05087}
}

@misc{favero2024enhancingcriticalthinkingeducation,
  title = {Enhancing Critical Thinking in Education by means of a Socratic Chatbot},
  author = {Lucile Favero and Juan Antonio Pérez-Ortiz and Tanja Käser and Nuria Oliver},
  year = {2024},
  eprint = {2409.05511},
  archivePrefix = {arXiv},
  primaryClass = {cs.HC},
  url = {https://arxiv.org/abs/2409.05511}
}

@misc{zhang2024splsocraticplaygroundlearning,
  title = {SPL: A Socratic Playground for Learning Powered by Large Language Model},
  author = {Liang Zhang and Jionghao Lin and Ziyi Kuang and Sheng Xu and Xiangen Hu},
  year = {2024},
  eprint = {2406.13919},
  archivePrefix = {arXiv},
  primaryClass = {cs.AI},
  url = {https://arxiv.org/abs/2406.13919}
}

@inproceedings{liu2024socraticLM,
  author = {Liu, Jiayu and Huang, Zhenya and Xiao, Tong and Sha, Jing and Wu, Jinze and Liu, Qi and Wang, Shijin and Chen, Enhong},
  title = {SocraticLM: exploring socratic personalized teaching with large language models},
  year = {2024},
  isbn = {9798331314385},
  publisher = {Curran Associates Inc.},
  address = {Red Hook, NY, USA},
  booktitle = {Proceedings of the 38th International Conference on Neural Information Processing Systems},
  articleno = {2721},
  numpages = {29},
  location = {Vancouver, BC, Canada},
  series = {NIPS '24}
}

@inproceedings{kochmar-etal-2025-findings,
  title = {Findings of the {BEA} 2025 Shared Task on Pedagogical Ability Assessment of {AI}-powered Tutors},
  author = {Kochmar, Ekaterina and Maurya, Kaushal and Petukhova, Kseniia and Srivatsa, KV Aditya and Tack, Anaïs and Vasselli, Justin},
  editor = {Kochmar, Ekaterina and Alhafni, Bashar and Bexte, Marie and Burstein, Jill and Horbach, Andrea and Laarmann-Quante, Ronja and Tack, Anaïs and Yaneva, Victoria and Yuan, Zheng},
  booktitle = {Proceedings of the 20th Workshop on Innovative Use of NLP for Building Educational Applications (BEA 2025)},
  month = jul,
  year = {2025},
  address = {Vienna, Austria},
  publisher = {Association for Computational Linguistics},
  url = {https://aclanthology.org/2025.bea-1.77/},
  doi = {10.18653/v1/2025.bea-1.77},
  pages = {1011--1033},
  ISBN = {979-8-89176-270-1}
}

@misc{jiang2025sidbenchmarkingguidedinstruction,
  title = {SID: Benchmarking Guided Instruction Capabilities in STEM Education with a Socratic Interdisciplinary Dialogues Dataset},
  author = {Mei Jiang and Houping Yue and Bingdong Li and Hao Hao and Ying Qian and Bo Jiang and Aimin Zhou},
  year = {2025},
  eprint = {2508.04563},
  archivePrefix = {arXiv},
  primaryClass = {cs.AI},
  url = {https://arxiv.org/abs/2508.04563}
}

@article{cowie1999model,
  author = {Cowie, B. and Bell, B.},
  title = {A model of formative assessment in science education},
  journal = {Assessment in Education: Principles, Policy \& Practice},
  volume = {6},
  number = {1},
  pages = {101--116},
  year = {1999}
}

@article{black2009developing,
  author = {Black, P. and Wiliam, D.},
  title = {Developing the theory of formative assessment},
  journal = {Educational Assessment, Evaluation and Accountability (formerly: Journal of Personnel Evaluation in Education)},
  volume = {21},
  pages = {5--31},
  year = {2009}
}

@misc{grattafiori2024llama3herdmodels,
      title={The Llama 3 Herd of Models}, 
      author={Aaron Grattafiori and Abhimanyu Dubey and Abhinav Jauhri and Abhinav Pandey and Abhishek Kadian and Ahmad Al-Dahle and Aiesha Letman and Akhil Mathur and Alan Schelten and Alex Vaughan and Amy Yang and Angela Fan and Anirudh Goyal and Anthony Hartshorn and Aobo Yang and Archi Mitra and Archie Sravankumar and Artem Korenev and Arthur Hinsvark and Arun Rao and Aston Zhang and Aurelien Rodriguez and Austen Gregerson and Ava Spataru and Baptiste Roziere and Bethany Biron and Binh Tang and Bobbie Chern and Charlotte Caucheteux and Chaya Nayak and Chloe Bi and Chris Marra and Chris McConnell and Christian Keller and Christophe Touret and Chunyang Wu and Corinne Wong and Cristian Canton Ferrer and Cyrus Nikolaidis and Damien Allonsius and Daniel Song and Danielle Pintz and Danny Livshits and Danny Wyatt and David Esiobu and Dhruv Choudhary and Dhruv Mahajan and Diego Garcia-Olano and Diego Perino and Dieuwke Hupkes and Egor Lakomkin and Ehab AlBadawy and Elina Lobanova and Emily Dinan and Eric Michael Smith and Filip Radenovic and Francisco Guzmán and Frank Zhang and Gabriel Synnaeve and Gabrielle Lee and Georgia Lewis Anderson and Govind Thattai and Graeme Nail and Gregoire Mialon and Guan Pang and Guillem Cucurell and Hailey Nguyen and Hannah Korevaar and Hu Xu and Hugo Touvron and Iliyan Zarov and Imanol Arrieta Ibarra and Isabel Kloumann and Ishan Misra and Ivan Evtimov and Jack Zhang and Jade Copet and Jaewon Lee and Jan Geffert and Jana Vranes and Jason Park and Jay Mahadeokar and Jeet Shah and Jelmer van der Linde and Jennifer Billock and Jenny Hong and Jenya Lee and Jeremy Fu and Jianfeng Chi and Jianyu Huang and Jiawen Liu and Jie Wang and Jiecao Yu and Joanna Bitton and Joe Spisak and Jongsoo Park and Joseph Rocca and Joshua Johnstun and Joshua Saxe and Junteng Jia and Kalyan Vasuden Alwala and Karthik Prasad and Kartikeya Upasani and Kate Plawiak and Ke Li and Kenneth Heafield and Kevin Stone and Khalid El-Arini and Krithika Iyer and Kshitiz Malik and Kuenley Chiu and Kunal Bhalla and Kushal Lakhotia and Lauren Rantala-Yeary and Laurens van der Maaten and Lawrence Chen and Liang Tan and Liz Jenkins and Louis Martin and Lovish Madaan and Lubo Malo and Lukas Blecher and Lukas Landzaat and Luke de Oliveira and Madeline Muzzi and Mahesh Pasupuleti and Mannat Singh and Manohar Paluri and Marcin Kardas and Maria Tsimpoukelli and Mathew Oldham and Mathieu Rita and Maya Pavlova and Melanie Kambadur and Mike Lewis and Min Si and Mitesh Kumar Singh and Mona Hassan and Naman Goyal and Narjes Torabi and Nikolay Bashlykov and Nikolay Bogoychev and Niladri Chatterji and Ning Zhang and Olivier Duchenne and Onur Çelebi and Patrick Alrassy and Pengchuan Zhang and Pengwei Li and Petar Vasic and Peter Weng and Prajjwal Bhargava and Pratik Dubal and Praveen Krishnan and Punit Singh Koura and Puxin Xu and Qing He and Qingxiao Dong and Ragavan Srinivasan and Raj Ganapathy and Ramon Calderer and Ricardo Silveira Cabral and Robert Stojnic and Roberta Raileanu and Rohan Maheswari and Rohit Girdhar and Rohit Patel and Romain Sauvestre and Ronnie Polidoro and Roshan Sumbaly and Ross Taylor and Ruan Silva and Rui Hou and Rui Wang and Saghar Hosseini and Sahana Chennabasappa and Sanjay Singh and Sean Bell and Seohyun Sonia Kim and Sergey Edunov and Shaoliang Nie and Sharan Narang and Sharath Raparthy and Sheng Shen and Shengye Wan and Shruti Bhosale and Shun Zhang and Simon Vandenhende and Soumya Batra and Spencer Whitman and Sten Sootla and Stephane Collot and Suchin Gururangan and Sydney Borodinsky and Tamar Herman and Tara Fowler and Tarek Sheasha and Thomas Georgiou and Thomas Scialom and Tobias Speckbacher and Todor Mihaylov and Tong Xiao and Ujjwal Karn and Vedanuj Goswami and Vibhor Gupta and Vignesh Ramanathan and Viktor Kerkez and Vincent Gonguet and Virginie Do and Vish Vogeti and Vítor Albiero and Vladan Petrovic and Weiwei Chu and Wenhan Xiong and Wenyin Fu and Whitney Meers and Xavier Martinet and Xiaodong Wang and Xiaofang Wang and Xiaoqing Ellen Tan and Xide Xia and Xinfeng Xie and Xuchao Jia and Xuewei Wang and Yaelle Goldschlag and Yashesh Gaur and Yasmine Babaei and Yi Wen and Yiwen Song and Yuchen Zhang and Yue Li and Yuning Mao and Zacharie Delpierre Coudert and Zheng Yan and Zhengxing Chen and Zoe Papakipos and Aaditya Singh and Aayushi Srivastava and Abha Jain and Adam Kelsey and Adam Shajnfeld and Adithya Gangidi and Adolfo Victoria and Ahuva Goldstand and Ajay Menon and Ajay Sharma and Alex Boesenberg and Alexei Baevski and Allie Feinstein and Amanda Kallet and Amit Sangani and Amos Teo and Anam Yunus and Andrei Lupu and Andres Alvarado and Andrew Caples and Andrew Gu and Andrew Ho and Andrew Poulton and Andrew Ryan and Ankit Ramchandani and Annie Dong and Annie Franco and Anuj Goyal and Aparajita Saraf and Arkabandhu Chowdhury and Ashley Gabriel and Ashwin Bharambe and Assaf Eisenman and Azadeh Yazdan and Beau James and Ben Maurer and Benjamin Leonhardi and Bernie Huang and Beth Loyd and Beto De Paola and Bhargavi Paranjape and Bing Liu and Bo Wu and Boyu Ni and Braden Hancock and Bram Wasti and Brandon Spence and Brani Stojkovic and Brian Gamido and Britt Montalvo and Carl Parker and Carly Burton and Catalina Mejia and Ce Liu and Changhan Wang and Changkyu Kim and Chao Zhou and Chester Hu and Ching-Hsiang Chu and Chris Cai and Chris Tindal and Christoph Feichtenhofer and Cynthia Gao and Damon Civin and Dana Beaty and Daniel Kreymer and Daniel Li and David Adkins and David Xu and Davide Testuggine and Delia David and Devi Parikh and Diana Liskovich and Didem Foss and Dingkang Wang and Duc Le and Dustin Holland and Edward Dowling and Eissa Jamil and Elaine Montgomery and Eleonora Presani and Emily Hahn and Emily Wood and Eric-Tuan Le and Erik Brinkman and Esteban Arcaute and Evan Dunbar and Evan Smothers and Fei Sun and Felix Kreuk and Feng Tian and Filippos Kokkinos and Firat Ozgenel and Francesco Caggioni and Frank Kanayet and Frank Seide and Gabriela Medina Florez and Gabriella Schwarz and Gada Badeer and Georgia Swee and Gil Halpern and Grant Herman and Grigory Sizov and Guangyi and Zhang and Guna Lakshminarayanan and Hakan Inan and Hamid Shojanazeri and Han Zou and Hannah Wang and Hanwen Zha and Haroun Habeeb and Harrison Rudolph and Helen Suk and Henry Aspegren and Hunter Goldman and Hongyuan Zhan and Ibrahim Damlaj and Igor Molybog and Igor Tufanov and Ilias Leontiadis and Irina-Elena Veliche and Itai Gat and Jake Weissman and James Geboski and James Kohli and Janice Lam and Japhet Asher and Jean-Baptiste Gaya and Jeff Marcus and Jeff Tang and Jennifer Chan and Jenny Zhen and Jeremy Reizenstein and Jeremy Teboul and Jessica Zhong and Jian Jin and Jingyi Yang and Joe Cummings and Jon Carvill and Jon Shepard and Jonathan McPhie and Jonathan Torres and Josh Ginsburg and Junjie Wang and Kai Wu and Kam Hou U and Karan Saxena and Kartikay Khandelwal and Katayoun Zand and Kathy Matosich and Kaushik Veeraraghavan and Kelly Michelena and Keqian Li and Kiran Jagadeesh and Kun Huang and Kunal Chawla and Kyle Huang and Lailin Chen and Lakshya Garg and Lavender A and Leandro Silva and Lee Bell and Lei Zhang and Liangpeng Guo and Licheng Yu and Liron Moshkovich and Luca Wehrstedt and Madian Khabsa and Manav Avalani and Manish Bhatt and Martynas Mankus and Matan Hasson and Matthew Lennie and Matthias Reso and Maxim Groshev and Maxim Naumov and Maya Lathi and Meghan Keneally and Miao Liu and Michael L. Seltzer and Michal Valko and Michelle Restrepo and Mihir Patel and Mik Vyatskov and Mikayel Samvelyan and Mike Clark and Mike Macey and Mike Wang and Miquel Jubert Hermoso and Mo Metanat and Mohammad Rastegari and Munish Bansal and Nandhini Santhanam and Natascha Parks and Natasha White and Navyata Bawa and Nayan Singhal and Nick Egebo and Nicolas Usunier and Nikhil Mehta and Nikolay Pavlovich Laptev and Ning Dong and Norman Cheng and Oleg Chernoguz and Olivia Hart and Omkar Salpekar and Ozlem Kalinli and Parkin Kent and Parth Parekh and Paul Saab and Pavan Balaji and Pedro Rittner and Philip Bontrager and Pierre Roux and Piotr Dollar and Polina Zvyagina and Prashant Ratanchandani and Pritish Yuvraj and Qian Liang and Rachad Alao and Rachel Rodriguez and Rafi Ayub and Raghotham Murthy and Raghu Nayani and Rahul Mitra and Rangaprabhu Parthasarathy and Raymond Li and Rebekkah Hogan and Robin Battey and Rocky Wang and Russ Howes and Ruty Rinott and Sachin Mehta and Sachin Siby and Sai Jayesh Bondu and Samyak Datta and Sara Chugh and Sara Hunt and Sargun Dhillon and Sasha Sidorov and Satadru Pan and Saurabh Mahajan and Saurabh Verma and Seiji Yamamoto and Sharadh Ramaswamy and Shaun Lindsay and Shaun Lindsay and Sheng Feng and Shenghao Lin and Shengxin Cindy Zha and Shishir Patil and Shiva Shankar and Shuqiang Zhang and Shuqiang Zhang and Sinong Wang and Sneha Agarwal and Soji Sajuyigbe and Soumith Chintala and Stephanie Max and Stephen Chen and Steve Kehoe and Steve Satterfield and Sudarshan Govindaprasad and Sumit Gupta and Summer Deng and Sungmin Cho and Sunny Virk and Suraj Subramanian and Sy Choudhury and Sydney Goldman and Tal Remez and Tamar Glaser and Tamara Best and Thilo Koehler and Thomas Robinson and Tianhe Li and Tianjun Zhang and Tim Matthews and Timothy Chou and Tzook Shaked and Varun Vontimitta and Victoria Ajayi and Victoria Montanez and Vijai Mohan and Vinay Satish Kumar and Vishal Mangla and Vlad Ionescu and Vlad Poenaru and Vlad Tiberiu Mihailescu and Vladimir Ivanov and Wei Li and Wenchen Wang and Wenwen Jiang and Wes Bouaziz and Will Constable and Xiaocheng Tang and Xiaojian Wu and Xiaolan Wang and Xilun Wu and Xinbo Gao and Yaniv Kleinman and Yanjun Chen and Ye Hu and Ye Jia and Ye Qi and Yenda Li and Yilin Zhang and Ying Zhang and Yossi Adi and Youngjin Nam and Yu and Wang and Yu Zhao and Yuchen Hao and Yundi Qian and Yunlu Li and Yuzi He and Zach Rait and Zachary DeVito and Zef Rosnbrick and Zhaoduo Wen and Zhenyu Yang and Zhiwei Zhao and Zhiyu Ma},
      year={2024},
      eprint={2407.21783},
      archivePrefix={arXiv},
      primaryClass={cs.AI},
      url={https://arxiv.org/abs/2407.21783}, 
}

@misc{liu2019robertarobustlyoptimizedbert,
      title={RoBERTa: A Robustly Optimized BERT Pretraining Approach}, 
      author={Yinhan Liu and Myle Ott and Naman Goyal and Jingfei Du and Mandar Joshi and Danqi Chen and Omer Levy and Mike Lewis and Luke Zettlemoyer and Veselin Stoyanov},
      year={2019},
      eprint={1907.11692},
      archivePrefix={arXiv},
      primaryClass={cs.CL},
      url={https://arxiv.org/abs/1907.11692}, 
}

@misc{qwen2025qwen25technicalreport,
      title={Qwen2.5 Technical Report}, 
      author={Qwen and : and An Yang and Baosong Yang and Beichen Zhang and Binyuan Hui and Bo Zheng and Bowen Yu and Chengyuan Li and Dayiheng Liu and Fei Huang and Haoran Wei and Huan Lin and Jian Yang and Jianhong Tu and Jianwei Zhang and Jianxin Yang and Jiaxi Yang and Jingren Zhou and Junyang Lin and Kai Dang and Keming Lu and Keqin Bao and Kexin Yang and Le Yu and Mei Li and Mingfeng Xue and Pei Zhang and Qin Zhu and Rui Men and Runji Lin and Tianhao Li and Tianyi Tang and Tingyu Xia and Xingzhang Ren and Xuancheng Ren and Yang Fan and Yang Su and Yichang Zhang and Yu Wan and Yuqiong Liu and Zeyu Cui and Zhenru Zhang and Zihan Qiu},
      year={2025},
      eprint={2412.15115},
      archivePrefix={arXiv},
      primaryClass={cs.CL},
      url={https://arxiv.org/abs/2412.15115}, 
}

@misc{jiang2023mistral7b,
      title={Mistral 7B}, 
      author={Albert Q. Jiang and Alexandre Sablayrolles and Arthur Mensch and Chris Bamford and Devendra Singh Chaplot and Diego de las Casas and Florian Bressand and Gianna Lengyel and Guillaume Lample and Lucile Saulnier and Lélio Renard Lavaud and Marie-Anne Lachaux and Pierre Stock and Teven Le Scao and Thibaut Lavril and Thomas Wang and Timothée Lacroix and William El Sayed},
      year={2023},
      eprint={2310.06825},
      archivePrefix={arXiv},
      primaryClass={cs.CL},
      url={https://arxiv.org/abs/2310.06825}, 
}

@misc{dettmers2023qloraefficientfinetuningquantized,
      title={QLoRA: Efficient Finetuning of Quantized LLMs}, 
      author={Tim Dettmers and Artidoro Pagnoni and Ari Holtzman and Luke Zettlemoyer},
      year={2023},
      eprint={2305.14314},
      archivePrefix={arXiv},
      primaryClass={cs.LG},
      url={https://arxiv.org/abs/2305.14314}, 
}

@misc{micikevicius2018mixedprecisiontraining,
      title={Mixed Precision Training}, 
      author={Paulius Micikevicius and Sharan Narang and Jonah Alben and Gregory Diamos and Erich Elsen and David Garcia and Boris Ginsburg and Michael Houston and Oleksii Kuchaiev and Ganesh Venkatesh and Hao Wu},
      year={2018},
      eprint={1710.03740},
      archivePrefix={arXiv},
      primaryClass={cs.AI},
      url={https://arxiv.org/abs/1710.03740}, 
}

@inbook{Scardamalia_Bereiter_2014, place={Cambridge}, series={Cambridge Handbooks in Psychology}, title={Knowledge Building and Knowledge Creation: Theory, Pedagogy, and Technology}, booktitle={The Cambridge Handbook of the Learning Sciences}, publisher={Cambridge University Press}, author={Scardamalia, Marlene and Bereiter, Carl}, editor={Sawyer, R. KeithEditor}, year={2014}, pages={397–417}, collection={Cambridge Handbooks in Psychology}}

@misc{khanmigo,
  title        = {Khanmigo},
  author       = {{Khan Academy}},
  year         = {2026},
  howpublished = {\url{https://www.khanmigo.ai/}},
  note         = {GPT-4-powered AI tutoring tool for learners, teachers, and parents; accessed 2026-03-29},
}

@article{https://doi.org/10.1207/s15516709cog2504_1,
author = {Chi, Michelene T.H. and Siler, Stephanie A. and Jeong, Heisawn and Yamauchi, Takashi and Hausmann, Robert G.},
title = {Learning from human tutoring},
journal = {Cognitive Science},
volume = {25},
number = {4},
pages = {471-533},
doi = {https://doi.org/10.1207/s15516709cog2504\_1},
url = {https://onlinelibrary.wiley.com/doi/abs/10.1207/s15516709cog2504_1},
eprint = {https://onlinelibrary.wiley.com/doi/pdf/10.1207/s15516709cog2504_1},
year = {2001}
}

@misc{hardy2026knowledgewisdommeasuringmisalignment,
      title={Knowledge without Wisdom: Measuring Misalignment between LLMs and Intended Impact}, 
      author={Michael Hardy and Yunsung Kim},
      year={2026},
      eprint={2603.00883},
      archivePrefix={arXiv},
      primaryClass={cs.LG},
      url={https://arxiv.org/abs/2603.00883}, 
}

@inproceedings{demszky-hill-2023-ncte,
    title = "The {NCTE} Transcripts: A Dataset of Elementary Math Classroom Transcripts",
    author = "Demszky, Dorottya  and
      Hill, Heather",
    editor = {Kochmar, Ekaterina  and
      Burstein, Jill  and
      Horbach, Andrea  and
      Laarmann-Quante, Ronja  and
      Madnani, Nitin  and
      Tack, Ana{\"i}s  and
      Yaneva, Victoria  and
      Yuan, Zheng  and
      Zesch, Torsten},
    booktitle = "Proceedings of the 18th Workshop on Innovative Use of NLP for Building Educational Applications (BEA 2023)",
    month = jul,
    year = "2023",
    address = "Toronto, Canada",
    publisher = "Association for Computational Linguistics",
    url = "https://aclanthology.org/2023.bea-1.44/",
    doi = "10.18653/v1/2023.bea-1.44",
    pages = "528--538"
}

@misc{macina2023mathdialdialoguetutoringdataset,
      title={MathDial: A Dialogue Tutoring Dataset with Rich Pedagogical Properties Grounded in Math Reasoning Problems}, 
      author={Jakub Macina and Nico Daheim and Sankalan Pal Chowdhury and Tanmay Sinha and Manu Kapur and Iryna Gurevych and Mrinmaya Sachan},
      year={2023},
      eprint={2305.14536},
      archivePrefix={arXiv},
      primaryClass={cs.CL},
      url={https://arxiv.org/abs/2305.14536}, 
}

@misc{beltagy2020longformerlongdocumenttransformer,
      title={Longformer: The Long-Document Transformer}, 
      author={Iz Beltagy and Matthew E. Peters and Arman Cohan},
      year={2020},
      eprint={2004.05150},
      archivePrefix={arXiv},
      primaryClass={cs.CL},
      url={https://arxiv.org/abs/2004.05150}, 
}

@article{pianta2008classroom,
  author = {Pianta, Robert C. and Belsky, Jay and Vandergrift, Nathan 
            and Houts, Renate and Morrison, Fred J.},
  title = {Classroom Effects on Children's Achievement Trajectories 
           in Elementary School},
  journal = {American Educational Research Journal},
  volume = {45},
  number = {2},
  pages = {365--397},
  year = {2008}
}

@article{hill2008mathematical,
  author = {Hill, Heather C. and Blunk, Merrie L. and Charalambous, 
            Charalambos Y. and Lewis, Jennifer M. and Phelps, Geoffrey C. 
            and Sleep, Laurie and Ball, Deborah Loewenberg},
  title = {Mathematical Knowledge for Teaching and the Mathematical 
           Quality of Instruction: An Exploratory Study},
  journal = {Cognition and Instruction},
  volume = {26},
  number = {4},
  pages = {430--511},
  year = {2008}
}

@misc{eriksson2025trustaibenchmarksinterdisciplinary,
      title={Can We Trust AI Benchmarks? An Interdisciplinary Review of Current Issues in AI Evaluation}, 
      author={Maria Eriksson and Erasmo Purificato and Arman Noroozian and Joao Vinagre and Guillaume Chaslot and Emilia Gomez and David Fernandez-Llorca},
      year={2025},
      eprint={2502.06559},
      archivePrefix={arXiv},
      primaryClass={cs.AI},
      url={https://arxiv.org/abs/2502.06559}, 
}

@inproceedings{10.1145/3442188.3445922,
author = {Bender, Emily M. and Gebru, Timnit and McMillan-Major, Angelina and Shmitchell, Shmargaret},
title = {On the Dangers of Stochastic Parrots: Can Language Models Be Too Big?},
year = {2021},
isbn = {9781450383097},
publisher = {Association for Computing Machinery},
address = {New York, NY, USA},
url = {https://doi.org/10.1145/3442188.3445922},
doi = {10.1145/3442188.3445922},
booktitle = {Proceedings of the 2021 ACM Conference on Fairness, Accountability, and Transparency},
pages = {610–623},
numpages = {14},
location = {Virtual Event, Canada},
series = {FAccT '21}
}
\bibliographystyle{colm2026_conference}
\appendix



\scriptsize
\setlength{\parskip}{0.5em}
\raggedright
\section{KB Dimensions (with examples)}
\label{app:kb_dimensions}

\noindent\textbf{Scaffolding}

\textit{Cognitive Engagement}

\textit{Definition:} Offering targeted support that helps students tackle challenges, gradually reducing guidance as their confidence and ability grow. Scaffolding requires teachers to recognize where a student is in their learning and provide just enough help to move them forward. As students develop independence, the teacher steps back, allowing them to take greater ownership. It's about balancing support with autonomy.

\textit{Initial Question:} ``What human activities could potentially increase global temperatures? I'm researching climate change for a science project.''

\textit{Excerpt:}

\vspace{0.2em}
\noindent\colorbox{black!5}{\begin{minipage}{\dimexpr\linewidth-2\fboxsep}
\vspace{0.3em}
\ttfamily\tiny\raggedright
...\\[0.05cm]
\textbf{Teacher:} That is a good example. Burning gasoline in cars release a major greenhouse gas, CO2(carbon dioxide). \textbf{Let's expand this idea, can you think further of other things or activities that can fuel or create pollution which create more greenhouse gases?}\\[0.05cm]
\textbf{Student:} Okay, so cars... What about factories? They make smoke, and that's gotta be bad, right? Is smoke a greenhouse gas too?\\[0.05cm]
\textbf{Teacher:} \textbf{You are right, Jamie.} Factories do give off smoke, but not all smoke itself is a greenhouse gas. \textbf{Another question, what do factories usually burn to get energy to run their machines?}\\[0.05cm]
\textbf{Student:} Huh. Good question. I guess they burn stuff like coal or oil or something? Is what makes greenhouse gases?\\[0.05cm]
\textbf{Teacher:} \textbf{That is it!} Factories often burn coal, oil or natural gas and these are called fossil fuels. \textbf{Now let's focus things we do to our land itself that might affect, can you mention any?}\\[0.05cm]
\textbf{...}\\[0.3em]
\end{minipage}}

\vspace{0.3em}
\textit{Why:} The teacher provides help to move Jamie's ideas forward by validating their responses and posing targeted guiding questions that break down the complex topic into manageable steps (cars → factories → land use).

\vspace{1em}
\noindent\rule{\linewidth}{0.4pt}
\vspace{0.5em}

\noindent\textbf{Critical Thinking}

\textit{Cognitive Engagement}

\textit{Definition:} Helping students analyze evidence, question assumptions, and draw logical conclusions to build deeper understanding. This dimension is about guiding students to go beyond surface-level answers. Teachers can encourage learners to weigh different pieces of information, identify inconsistencies or gaps in reasoning, and construct explanations based on sound logic and evidence. It's about cultivating a habit of intellectual curiosity and disciplined reasoning.

\textit{Initial Question:} ``What's the best evidence that Earth's continents were once positioned differently than they are today?''

\textit{Excerpt:}

\vspace{0.2em}
\noindent\colorbox{black!5}{\begin{minipage}{\dimexpr\linewidth-2\fboxsep}
\vspace{0.3em}
\ttfamily\tiny\raggedright
...\\[0.05cm]
\textbf{Teacher:} Yes, the continents really do look like pieces of a giant jigsaw puzzle! \textbf{Why do you think that's important for figuring out if they moved?}\\[0.05cm]
\textbf{Student:} Well, if they fit together, they were probably, like, together, right? Before they drifted apart or something?\\[0.05cm]
\textbf{Teacher:} That's right! And scientists have found even more clues. For example, fossils of tropical plants have been discovered in Antarctica. \textbf{Does that sound like something you'd expect in such a cold place today?}\\[0.05cm]
\textbf{Student:} Wait, what? Tropical plants in Antarctica? That's crazy! So, like, Antarctica used to be warm? Whoa.\\[0.05cm]
\textbf{Teacher:} It wild, right! That discovery tells us Antarctica was once closer to the equator---where it was warm enough for those plants to grow. \textbf{So if a whole continent moved that far, what does that say about Earth's surface over time?}\\[0.05cm]
\textbf{Student:} So, like, the Earth's surface... changes? A lot? Like, way more than I thought. It's not just the water that moves, the land does too?! Mind blown.\\[0.05cm]
\textbf{Teacher:} Exactly! The land itself moves over millions of years because of something called plate tectonics. \textbf{What other clues might help prove that continents used to be in different spots?}\\[0.05cm]
...\\[0.3em]
\end{minipage}}

\vspace{0.3em}
\textit{Why:} The teacher guides Jamie to analyze evidence and draw logical conclusions by prompting them to evaluate multiple types of proof (fossils, continental fit, plant locations), question expectations (``Does that sound like something you'd expect...?''), and synthesize clues into a coherent argument for continental drift.

\normalsize

\noindent\rule{\linewidth}{0.4pt}

\scriptsize
\setlength{\parskip}{0.5em}
\raggedright

\noindent\textbf{Generative Questioning}

\textit{Cognitive Engagement}

\textit{Definition:} Posing open-ended, thought-provoking questions that guide students to connect, explore, and extend their ideas. In this dimension, the teacher uses questions not just to test knowledge, but to inspire deeper inquiry. By framing questions that don't have one right answer, teachers create space for students to speculate, wonder, and generate new perspectives. It's about activating the learner's natural curiosity and helping them construct meaning.

\textit{Initial Question:} ``What type of galaxy is our Milky Way, and where is our solar system located within it?''

\textit{Excerpt:}

\vspace{0.2em}
\noindent\colorbox{black!5}{\begin{minipage}{\dimexpr\linewidth-2\fboxsep}
\vspace{0.3em}
\ttfamily\tiny\raggedright
...\\[0.05cm]
\textbf{Teacher:} that is a good observation jamie, you are very right ,many galaxies, inncluding even our own milky way we call those spiral galaxies.now that you have asked if there are other shapes ,\textbf{What other ways could you imagine a galaxy looking if it was not spiral? and how might knowing the shape of our galaxy help us figure out where our solar system is located within?}\\[0.05cm]
\textbf{Student:} Hmm... maybe like a blob? Or a big oval? I guess if it's swirly, maybe we're like... on one of the swirls? But I don't really get how knowing the shape helps.\\[0.05cm]
\textbf{Teacher:} you are on the right track, those are also types of galaxies,we call them elliptical and irregular galaxies,,now,thinking about your idea of being on one of the swirls.\textbf{what does that suggest about how the stars and other parts of galaxy are organized?and if we are on swirl how might that affect our view of the rest of the galaxy from earth?}\\[0.05cm]
\textbf{Student:} Organized... like, all going the same direction? So if we're on a swirl, we'd see more stars in the direction of the swirl... I guess? Is that right?\\[0.05cm]
\textbf{Teacher:} that is a very insightul thought jamie,you are just thinking from my perspective,if the stars and the gas in a spiral galaxy are indeed organized into those swirls or arms and say maybe we are in one of them \textbf{what might that imply about how easy or difficult is it to see the entire shape of our own galaxy from earth?what might block our view}\\[0.05cm]
...\\[0.3em]
\end{minipage}}

\vspace{0.3em}
\textit{Why:} The teacher uses ``open-ended, thought-provoking questions'' that don't have single right answers (``What other ways could you imagine a galaxy looking?'', ``how might that affect our view?''), prompting Jamie to speculate, and make connections between ideas.

\vspace{1em}
\noindent\rule{\linewidth}{0.4pt}
\vspace{0.5em}

\noindent\textbf{Problem-based Learning}

\textit{Cognitive Engagement}

\textit{Definition:} Presenting real-world problems that prompt students to explore, inquire, and construct understanding through active problem solving. This dimension emphasizes learning by doing. Teachers act as facilitators who pose meaningful challenges and support students as they navigate complexity, test ideas, and apply concepts to new situations. It's about making learning relevant and transferable.

\textit{Initial Question:} ``How do ocean currents like the Gulf Stream affect the climate of nearby coastal areas?''

\textit{Excerpt:}

\vspace{0.2em}
\noindent\colorbox{black!5}{\begin{minipage}{\dimexpr\linewidth-2\fboxsep}
\vspace{0.3em}
\ttfamily\tiny\raggedright
...\\[0.05cm]
\textbf{Teacher:} Thats a really interesting observation! Lets think about it in real world scenario. \textbf{Imagine you're working with a team designing a climate exhibit for a science museum. One part of the exhibit compresses the temperature of London and St. John's Newfoundland, both at similar latitudes, but with very different winter climates. You need to explain to visitors how the Gulf Stream, a warm ocean current, actually helps keep Europe warmer. How does warm ocean waer actually transfer heat to the air, and how does that air affect land temperatures nearby? What would your exhibit need to show or demonstrate to make that understandable?}\\[0.05cm]
\textbf{Student:} Okay... so like, the water is warm, right? Does it, like, warm up the air above it? Then the wind blows that warm air onto the land? So like, maybe the exhibit could have a fan blowing over a tub of warm water and then onto a little fake beach or something? I dunno. It's still kinda weird. Like, my bathwater doesn't heat up my whole bathroom.\\[0.05cm]
\textbf{Teacher:} That's a really creative idea,using a fan and a tub to show how warm water can heat air and blow it towards land! You're onto something important , the ocean doesn't heat the land directly, it warms the air above it and winds carry the the warm air over the coast. \textbf{But now, here's your challenge, why do you think this works on such a large scale like with the Gulf Stream, but not in your bathroom? Whats different about the amount of water, the temperature or the air movement?}\\[0.05cm]
\textbf{...}\\[0.3em]
\end{minipage}}

\vspace{0.3em}
\textit{Why:} The teacher presents a real-world problem (designing a museum exhibit) that requires Jamie to construct understanding through active problem solving, making the abstract concept of ocean currents tangible.

\normalsize

\noindent\rule{\linewidth}{0.4pt}

\scriptsize
\setlength{\parskip}{0.5em}
\raggedright

\noindent\textbf{Ownership of Ideas}

\textit{Power Dynamics}

\textit{Definition:} Empowering students to take intellectual ownership by generating, developing, and refining their own ideas. Instead of just absorbing information, students are encouraged to shape their understanding through personal engagement. Teachers can foster this by validating student contributions, encouraging revision and elaboration, and making space for student-led inquiry.

\textit{Initial Question:} ``My teacher mentioned Venus has a greenhouse effect. Why is Venus so hot despite its thick cloud cover that blocks much of the Sun's light?''

\textit{Excerpt:}

\vspace{0.2em}
\noindent\colorbox{black!5}{\begin{minipage}{\dimexpr\linewidth-2\fboxsep}
\vspace{0.3em}
\ttfamily\tiny\raggedright
...\\[0.05cm]
\textbf{Teacher:} Great question, what do you think is going on? If the clouds block sunlight, why might Venus still heat up?\\[0.05cm]
\textbf{Student:} Hmm, I dunno. Maybe the clouds are like a blanket? Like how I get hot under my covers even though it's dark?\\[0.05cm]
\textbf{Teacher:} \textbf{That's a really smart comparison! So if the clouds act like a blanket, what do you think they might be on Venus?}\\[0.05cm]
\textbf{Student:} Oh! Are they trapping heat? Like, the sun's heat gets in, but then it can't get back out?\\[0.05cm]
\textbf{Teacher:} exactly good answer\\[0.05cm]
\textbf{Student:} Cool! So, even though the clouds block sunlight, the heat that get through gets stuck? That makes sense.\\[0.05cm]
\textbf{Teacher:} Exactly! \textbf{You're piecing it together really well.} What do you think happens to that trapped heat over time does it just stay the same, or build up?\\[0.05cm]
\textbf{Student:} I guess it builds up, right? ... Kinda like a car in the sun with the windows rolled up.\\[0.05cm]
\textbf{Teacher:} \textbf{You nailed it, Jamie! You took your own idea like getting warm under a blanket and built a solid science explanation from it.}\\[0.05cm]
\textbf{...}\\[0.3em]
\end{minipage}}

\vspace{0.3em}
\textit{Why:} The teacher empowers Jamie to generate and refine their own ideas by validating their original ``blanket'' analogy, encouraging them to extend it (``what do you think they might be on Venus?''), and explicitly recognizing how they transformed her personal comparison into a scientific explanation.

\vspace{1em}
\noindent\rule{\linewidth}{0.4pt}
\vspace{0.5em}

\noindent\textbf{Partisanship}

\textit{Power Dynamics}

\textit{Definition:} Encouraging students to take positions on issues while remaining open to and respectful of differing perspectives. This dimension supports students in developing informed opinions and expressing them confidently. Teachers help students navigate the tension between advocacy and openness, fostering civic reasoning and dialogue.

\textit{Initial Question:} ``What human activities contribute to ozone depletion in the upper atmosphere? Why is this concerning?''

\textit{Excerpt:}

\vspace{0.2em}
\noindent\colorbox{black!5}{\begin{minipage}{\dimexpr\linewidth-2\fboxsep}
\vspace{0.3em}
\ttfamily\tiny\raggedright
...\\[0.05cm]
\textbf{Teacher:} ...only certain human-made chemicals are the main culprits. For example, substances called chlorofluorocarbons (CFCs), halons, and a few others are especially damaging... \textbf{Now that you know this information, would you take the position that we should focus our efforts on banning or controlling these specific chemicals, rather than all pollutants, to protect the ozone layer? Why or why not? Try to defend your stance with a reason or two!}\\[0.05cm]
\textbf{Student:} Oh, okay. So, like, not chemicals are bad for the ozone layer, just some specific ones. So, yeah, I guess we should probably focus on, like, banning chemicals. It kinda makes sense to, you know, focus on the stuff that's actually doing the damage.\\[0.05cm]
\textbf{Teacher:} Exactly, Jamie! ...You're taking a clear position here, which is great! \textbf{Now, some people argue that banning these chemicals could hurt industries or make products more expensive. Others say that protecting the ozone layer should come first because it shields us from dangerous UV radiation. Which side do you lean toward, and why? Can you use any evidence or reasoning to back up your position?}\\[0.05cm]
\textbf{Student:} Hmm... that's kinda tricky. I guess... I mean, getting sunburned is bad, right? So maybe protecting the ozone layer is more important? Like, what if more people get skin cancer or something? That's worse than, like, stuff being more expensive, I guess.\\[0.05cm]
\textbf{Teacher:} You're making a thoughtful argument, Jamie! ...For example, if everyone agrees to remove ozone-depleting chemicals, it will prevent about 443 million cases of skin cancer... \textbf{Given this evidence, do you feel confident defending your position that protecting the ozone layer is more important than the potential economic drawbacks? Or do you see any reasons why someone might still disagree?}\\[0.05cm]
\textbf{...}\\[0.3em]
\end{minipage}}

\vspace{0.3em}
\textit{Why:} The teacher encourages Jamie to take positions on issues by explicitly asking them to choose and defend a stance, presenting opposing economic concerns, and prompting them to consider why others might disagree, fostering respectful dialogue.

\normalsize

\noindent\rule{\linewidth}{0.4pt}

\scriptsize
\setlength{\parskip}{0.5em}
\raggedright

\noindent\textbf{Persuasive Discourse}

\textit{Power Dynamics}

\textit{Definition:} Helping students articulate and defend their ideas using evidence, while engaging respectfully with alternative viewpoints. This dimension emphasizes respectful, evidence-based argumentation. Teachers encourage students to back up their opinions with data or reasoning, respond to counterpoints thoughtfully, and build collaborative conversations that value diverse perspectives.

\textit{Initial Question:} ``How did the planets in our solar system form? I've heard different theories but I'm not sure which is correct.''

\textit{Excerpt:}

\vspace{0.2em}
\noindent\colorbox{black!5}{\begin{minipage}{\dimexpr\linewidth-2\fboxsep}
\vspace{0.3em}
\ttfamily\tiny\raggedright
...\\[0.05cm]
\textbf{Teacher:} Yes, now lets think about the rocky and gassy types of planets. How might that relate to this phenomenon?\\[0.05cm]
\textbf{Student:} Okay, so... closer to the sun it's hotter, right? So maybe the lighter, gassy stuff got pushed farther away? And the heavier rocky stuff stayed closer? Is that why the planets close to the sun are rocky and the ones further away are gassy giants?\\[0.05cm]
\textbf{Teacher:} \textbf{Instead of asking me, go through your thought process with evidence and reasoning to convince me of your theory}\\[0.05cm]
\textbf{Student:} Okay, okay, so think about it. The sun is like a giant heater, right? So all the light stuff, like gas, would get blown away from the sun like a leaf in the wind. The heavy rocky stuff is harder to move, so it stays closer. That's why Mercury, Venus, Earth, and Mars are rocky 'cause they're close. And Jupiter, Saturn, Uranus, and Neptune are gassy giants 'cause they're far away. See? Makes sense!\\[0.05cm]
\textbf{Teacher:} \textbf{So let's put it all together. Let's pretend you have someone who is skeptical about this process. How would you explain it from start to finish?}\\[0.05cm]
\textbf{Student:} Okay, so imagine a HUGE cloud of dust and gas, right? Then BOOM! Something explodes... and it kinda mixes everything up. But...\\[0.05cm]
...\\[0.3em]
\end{minipage}}

\vspace{0.3em}
\textit{Why:} The teacher prompts Jamie to articulate and defend their ideas using evidence rather than seeking validation, then challenges them to address a skeptical audience, fostering evidence-based argumentation that requires them to construct a persuasive explanation.

\normalsize

\noindent\rule{\linewidth}{0.4pt}

\scriptsize
\setlength{\parskip}{0.5em}
\raggedright

\noindent\textbf{Cultural Analogies}

\textit{Cultural Responsiveness}

\textit{Definition:} Using culturally familiar examples to make complex or abstract ideas more accessible and meaningful to students. Teachers draw on metaphors, traditions, or real-life scenarios from students' communities to bridge gaps in understanding. This dimension highlights the value of contextualizing science in ways that resonate with diverse learners.

\textit{Initial Question:} ``Only about 50\% of solar energy directed toward Earth actually reaches the surface. What happens to the rest of it?''

\textit{Excerpt:}

\vspace{0.2em}
\noindent\colorbox{black!5}{\begin{minipage}{\dimexpr\linewidth-2\fboxsep}
\vspace{0.3em}
\ttfamily\tiny\raggedright
...\\[0.05cm]
\textbf{Teacher:} The Earth has many protective layers on it. \textbf{Can you think of something from your culture that acts as a protective layer? How does it work?}\\[0.05cm]
\textbf{Student:} Um...like my jacket? It keeps me warm by, like, blocking the wind and stuff. Is that what you mean?\\[0.05cm]
\textbf{Teacher:} Exactly! Just like your jacket blocks the wind, the ozone layer blocks UV (ultraviolet) radiation from the sun. What else protects you?\\[0.05cm]
\textbf{Student:} Oh, okay. Sunscreen? That blocks UV rays too, right? But what happens to the rest of the sun's energy, not just UV rays?\\[0.05cm]
\textbf{Teacher:} Great example. But suncreen doesn't just block - it reflects some of that energy also. Parts of the Earth can reflect solar energy back into space... \textbf{Can you think of anything from your culture that protects by absorbing?}\\[0.05cm]
\textbf{Student:} ...um... absorbing... Is it like a sponge? My mom uses those to clean up spills. They soak stuff up. Is the air like a giant sponge?\\[0.05cm]
\textbf{Teacher:} Fantastic! Yes - 30\% is immediatly reflected, while 20\% is absorbed by the atmospher - just like your mother's sponge...\\[0.05cm]
...\\[0.3em]
\end{minipage}}

\vspace{0.3em}
\textit{Why:} The teacher prompts Jamie to draw on familiar everyday objects (jacket, sunscreen, mother's sponge), shifting conversation about abstract atmospheric processes into concrete, culturally-grounded comparisons that resonate with their lived experience.

\vspace{1em}
\noindent\rule{\linewidth}{0.4pt}
\vspace{0.5em}

\noindent\textbf{Cultural Identity Exploration}

\textit{Cultural Responsiveness}

\textit{Definition:} Helping students reflect on how cultural backgrounds shape perspectives and interpretations of knowledge. Teachers can create opportunities for students to share personal stories, draw connections between their lived experiences and scientific ideas, and examine how culture influences thinking. This fosters inclusivity and relevance in the learning environment.

\textit{Initial Question:} ``What instruments do scientists use to detect and measure earthquakes? How do they work?''

\textit{Excerpt:}

\vspace{0.2em}
\noindent\colorbox{black!5}{\begin{minipage}{\dimexpr\linewidth-2\fboxsep}
\vspace{0.3em}
\ttfamily\tiny\raggedright
...\\[0.05cm]
\textbf{Teacher:} ...Inertia means that objects like to keep doing what they're doing --- if something is still, it wants to stay still..........\textbf{Also in your culture or family, have you ever heard any stories or explanations about what causes earthquakes?}\\[0.05cm]
\textbf{Student:} Oh, okay. Like when you're in a car and it stops suddenly, and you keep moving forward a little? So the heavy thing in the seismo...thingy, it's like that? But how does it it? Like, does it draw a picture or something? My grandpa says earthquakes are caused by giant catfish underground, but I don't think that's what you're looking for, haha.\\[0.05cm]
\textbf{Teacher:} Yes, that is exactly like the car example. ...And actually, \textbf{I love that story from your grandpa about the giant catfish. Stories like that are really important, they show how people have explained natural events for generations, long before modern science tools existed. Why do you think people in different cultures might create stories like that about earthquakes?}\\[0.05cm]
\textbf{Student:} ...Hmm, why stories about earthquakes? I guess 'cause they're scary and people wanted to understand why they happen, even if they didn't have seismo... graph... things.\\[0.05cm]
\textbf{Teacher:} Exactly, Jamie, you are getting it. ...You are also really thoughtful about why people tell those stories. Earthquakes can be scary, and long before modern science, people used stories to make sense of what was happening around them. \textbf{Do you think hearing those kinds of stories might change how people in different places feel about earthquakes, compared to just hearing the scientific explanation?}\\[0.05cm]
\textbf{...}\\[0.3em]
\end{minipage}}

\vspace{0.3em}
\textit{Why:} The teacher invites Jamie to share personal stories from their family (her grandpa's catfish story), validates how cultural backgrounds shape perspectives by affirming these stories as important ways people have historically understood earthquakes.

\normalsize

\noindent\rule{\linewidth}{0.4pt}

\scriptsize
\setlength{\parskip}{0.5em}
\raggedright

\noindent\textbf{Reflective Growth}

\textit{Metacognition}

\textit{Definition:} Helping students review past learning, recognize progress, and set new goals for continued improvement. Teachers can guide students in identifying milestones, celebrating achievements, and planning next steps. This fosters a growth mindset and long-term engagement.

\textit{Initial Question:} ``How can geologists determine the cooling rate of igneous intrusive rocks? What features do they look for?''

\textit{Excerpt:}

\vspace{0.2em}
\noindent\colorbox{black!5}{\begin{minipage}{\dimexpr\linewidth-2\fboxsep}
\vspace{0.3em}
\ttfamily\tiny\raggedright
...\\[0.05cm]
\textbf{Teacher:} yes way to think about this carefully. It forms small crystals or glass-nice deduction! Granite with large crystals indicates slow cooling deep underground while fine-grained basalt does lead up to believe it would be rapid cooling at the surface\\[0.05cm]
\textbf{Student:} Okay, that makes sense. So like, granite is the slow one, and basalt is the fast one. So if a rock has teeny tiny crystals, it cooled down super fast?\\[0.05cm]
\textbf{Teacher:} \textbf{I like how you can use this knowledge from our past units to make sense of this current unit. Great job. What can you note about your progress during this session so far?}\\[0.05cm]
\textbf{Student:} Well, at first I was totally lost, like I didn't even know to look at. But now I get that crystal size tells you how fast a rock cooled. Big crystals, slow cooling. Small crystals, fast cooling. I feel way better about it now!\\[0.05cm]
\textbf{Teacher:} \textbf{That's wonderful! What new goals would you like to set for yourself for continued improvement?}\\[0.05cm]
\textbf{Student:} Hmm... maybe I could try to remember the examples, like granite and basalt. And maybe try to think about things happen, like how the atoms need time to make big crystals.\\[0.05cm]
...\\[0.3em]
\end{minipage}}

\vspace{0.3em}
\textit{Why:} The teacher guides Jamie to review past learning, recognize progress, and set new goals by prompting explicit reflection on their learning journey (``What can you note about your progress?'') and future planning (``What new goals would you like to set?'').

\vspace{1em}
\noindent\rule{\linewidth}{0.4pt}
\vspace{0.5em}

\noindent\textbf{Strategic Thinking}

\textit{Metacognition}

\textit{Definition:} Teaching students to plan, monitor, and adapt learning strategies based on the task and their progress. Teachers support students in identifying goals, choosing appropriate methods, and reflecting on effectiveness. It's about helping learners become flexible, purposeful thinkers.

\textit{Initial Question:} ``Why does Saturn rotate so much faster than Earth? My astronomy homework says Saturn's day is only about 10 Earth hours long.''

\textit{Excerpt:}

\vspace{0.2em}
\noindent\colorbox{black!5}{\begin{minipage}{\dimexpr\linewidth-2\fboxsep}
\vspace{0.3em}
\ttfamily\tiny\raggedright
...\\[0.05cm]
\textbf{Teacher:} Great question, Jamie! \textbf{To figure this out, how might you start investigating why Saturn spins faster? What kind of information or strategies could help you understand this difference?}\\[0.05cm]
\textbf{Student:} Um... I guess I could look up how fast they both spin? Is it like, because Saturn is bigger? Or maybe made of different stuff?\\[0.05cm]
\textbf{Teacher:} Those are excellent ideas! Comparing their rotation speeds and considering factors like size and composition are great strategies. \textbf{Which one do you think would have the biggest effect on rotation speed? How could you find evidence to support your guess?}\\[0.05cm]
\textbf{Student:} Hmm... I think size. Like, if you spin a basketball and a baseball, the basketball is faster, right? But I dunno how to prove that for planets... Is there like, a planet website?\\[0.05cm]
\textbf{Teacher:} That's a good analogy, Jamie! To test your idea, you could look up reliable sources like NASA's website or astronomy databases to find data on Saturn's and Earth's rotation speeds and compositions. \textbf{After you gather that info, how might you compare it to see what influences rotation speed most?}\\[0.05cm]
\textbf{Student:} Oh, okay. So look up how fast they spin and what they're made of. Then... compare them? Like, make a chart or something? Is that what you mean?\\[0.05cm]
\textbf{Teacher:} \textbf{Exactly, Jamie! Creating a chart to compare their rotation speeds, sizes, and compositions is a smart way to organize your information and spot patterns. After making the chart, what would you look for to figure out why Saturn spins faster?}\\[0.05cm]
...\\[0.3em]
\end{minipage}}

\vspace{0.3em}
\textit{Why:} The teacher guides Jamie by prompting them to identify investigation methods, choose appropriate approaches (comparison chart), and reflect on which factors to prioritize, helping her plan own learning approach.

\normalsize
\noindent\rule{\linewidth}{0.4pt}


\scriptsize
\setlength{\parskip}{0.5em}
\raggedright

\noindent\textbf{Thinking Aloud}

\textit{Metacognition}

\textit{Definition:} Modeling reasoning and articulating decision-making processes to make scientific thinking visible to students. Teachers verbalize their own thought processes---asking questions, making predictions, revising ideas---to show how experts think. This helps demystify complex thinking and invites students to adopt similar strategies.

\textit{Initial Question:} ``What human activities could potentially increase global temperatures? I'm researching climate change for a science project.''

\textit{Excerpt:}

\vspace{0.2em}
\noindent\colorbox{black!5}{\begin{minipage}{\dimexpr\linewidth-2\fboxsep}
\vspace{0.3em}
\ttfamily\tiny\raggedright
...\\[0.05cm]
\textbf{Teacher:} Hmm... like, driving cars? Do those make the planet hotter? \textbf{Well, cars use gas, and when the gas burns, it makes smoke or gas that goes into the air. I think that gas is called carbon dioxide. I remember learning that carbon dioxide traps heat, kind of like a blanket around the Earth. So yeah... if a lot of people are driving cars every day, that probably puts a lot of that gas into the air and makes the planet warmer.}\\[0.05cm]
\textbf{Student:} Oh, okay. So, like, the more cars, the more blanket, the hotter it gets?\\[0.05cm]
\textbf{Teacher:} You're thinking in the right direction! Can you explain a little more about what that ``blanket'' is made of --- and where it comes from? What do you think happens when we add more and more of it?\\[0.05cm]
\textbf{Student:} It's that carbon di-oxide stuff, right? And it comes from burning stuff like gas in cars. If we keep adding more, it's like a thicker blanket, so it gets even hotter, I guess?\\[0.05cm]
\textbf{Teacher:} Yes! You're really putting it together! Now think about this: besides cars, what are some other things people do that might burn gas or other fuels and add more of that carbon dioxide ``blanket'' to the air?\\[0.05cm]
...\\[0.3em]
\end{minipage}}

\vspace{0.3em}
\textit{Why:} The teacher models reasoning by verbalizing their own thought process (``cars use gas...I think that gas is called carbon dioxide...I remember learning...''), making scientific thinking visible and inviting Jamie to adopt similar analytical strategies.

\noindent\rule{\linewidth}{0.4pt}

\noindent\textbf{Error Analysis}

\textit{Metacognition}

\textit{Definition:} Encouraging students to examine mistakes to uncover misunderstandings and refine future thinking. Instead of viewing errors as failures, students learn to see them as opportunities. Teachers can model how to diagnose a mistake, trace its cause, and revise understanding accordingly.

\textit{Initial Question:} ``I'm studying acid rain effects on different rock types. My textbook mentions an experiment where rocks in vinegar solutions showed different pH changes over time - some stayed at pH 3 while others reached pH 7. Why would this happen?''

\textit{Excerpt:}

\vspace{0.2em}
\noindent\colorbox{black!5}{\begin{minipage}{\dimexpr\linewidth-2\fboxsep}
\vspace{0.3em}
\ttfamily\tiny\raggedright
...\\[0.05cm]
\textbf{Teacher:} That's a very interesting observation, Jamie! It's a great question about how different rocks react to acid, which is key to understanding acid rain. \textbf{The textbook mentions some vinegar solutions staying at pH 3 and others reaching pH 7. When you see a result like that, what's your first thought about why the would behave so differently with different rocks? What could the rocks themselves be doing to the solution?}\\[0.05cm]
\textbf{Student:} Um... maybe some rocks are, like, soaking up the vinegar? Or, wait, are they, like, the vinegar somehow? I don't get how they could make it less acidic though.\\[0.05cm]
\textbf{Teacher:} You're definitely on the right track, Jamie, especially with ``changing the vinegar somehow''! \textbf{Rocks aren't really ``soaking up'' the vinegar like a sponge; instead, there's a chemical interaction happening between the rock and the acid. To make a solution less acidic (move from pH 3 towards pH 7), what kind of substance would you need to add to an acid? Think about the opposite of an acid in chemistry. What chemically with an acid to neutralize it?}\\[0.05cm]
\textbf{Student:} Oh, a base! Like baking soda and vinegar, right? That makes a volcano! So, are some rocks, like, basic? Is that why the vinegar changes?\\[0.05cm]
\textbf{Teacher:} You've hit on the exact chemical principle, Jamie! You're absolutely right -- it's just like baking soda and vinegar, which is a classic acid-base reaction. So yes, some rocks contain minerals that are basic (or alkaline). \textbf{Knowing this, how does the presence of these basic minerals within the rock explain why some vinegar solutions changed pH to 7, while others stayed at pH 3? What's actually happening in the solution when the acid encounters these basic parts of the rock?}\\[0.05cm]
\textbf{...}\\[0.3em]
\end{minipage}}

\vspace{0.3em}
\textit{Why:} The teacher helps Jamie examine their mistakes by addressing their initial misconception (rocks ``soaking up'' vinegar), redirecting them to examine what's chemically happening, and guiding them to trace the cause of the pH variation through understanding rock composition.

\normalsize

\noindent\rule{\linewidth}{0.4pt}

\scriptsize
\setlength{\parskip}{0.5em}
\raggedright

\noindent\textbf{Reflection}

\textit{Metacognition}

\textit{Definition:} Encouraging students to reflect on their thinking and learning processes to develop metacognitive awareness. Teachers can invite students to consider how they approached a task, what strategies they used, and what they might try differently next time. This builds learners' ability to self-monitor and adjust their thinking, which is essential for lifelong learning.

\textit{Initial Question:} ``What type of galaxy is our Milky Way, and where is our solar system located within it?''

\textit{Excerpt:}

\vspace{0.2em}
\noindent\colorbox{black!5}{\begin{minipage}{\dimexpr\linewidth-2\fboxsep}
\vspace{0.3em}
\ttfamily\tiny\raggedright
...\\[0.05cm]
\textbf{Teacher:} That's a good starting point! \textbf{Let's pause for a moment---when you try to learn something like this, what usually helps you remember details? Diagrams? Videos? Examples?}\\[0.05cm]
\textbf{Student:} Pictures definitely help. Like, if I see a diagram or something. Videos are okay, but sometimes they go too fast.\\[0.05cm]
\textbf{Teacher:} That's really helpful to know. Diagrams can slow things down and let you take it all in at your own pace. Let's build a mental picture together: imagine a giant pinwheel made of stars---that's our spiral galaxy, the Milky Way...\\[0.05cm]
\textbf{Student:} ...Okay, I think I get that. It's like, not in the dead center, but not at the very edge either. The pinwheel thing makes it easier to see, I guess.\\[0.05cm]
\textbf{Teacher:} Yes, you've got it! That's a really clear way to think about it. Now, quick question: \textbf{when you figured that out just now, what part helped it click for you---the visual of the pinwheel, thinking about the position, or something else?}\\[0.05cm]
\textbf{Student:} I think it was like, picturing it in my head. Like actually seeing the pinwheel and where our solar system would be.\\[0.05cm]
\textbf{Teacher:} That's great insight, Jamie. Knowing that visualizing helps you is really useful---not just for this, but for other science topics too. \textbf{Next time you run into something tricky, how might you use that strategy again to help yourself understand?}\\[0.05cm]
\textbf{...}\\[0.3em]
\end{minipage}}

\vspace{0.3em}
\textit{Why:} The teacher prompts Jamie to identify what helps them learn best and analyze which specific approach made the concept click, building awareness of their own cognitive patterns and how to transfer successful strategies to future challenges.

\normalsize

\noindent\rule{\linewidth}{0.4pt}

\scriptsize
\setlength{\parskip}{0.5em}
\raggedright

\noindent\textbf{Self-Reflection}

\textit{Metacognition}

\textit{Definition:} Prompting students to examine how their thoughts, emotions, and actions influence their learning. Teachers encourage introspection by asking students to assess their motivation, mindset, and habits. This helps learners become more intentional and resilient in their academic journey.

\textit{Initial Question:} ``Only about 50\% of solar energy directed toward Earth actually reaches the surface. What happens to the rest of it?''

\textit{Excerpt:}

\vspace{0.2em}
\noindent\colorbox{black!5}{\begin{minipage}{\dimexpr\linewidth-2\fboxsep}
\vspace{0.3em}
\ttfamily\tiny\raggedright
...\\[0.05cm]
\textbf{Teacher:} \textbf{Let's stop for a second: If a question is very confusing, what habitually happens? Do you try to guess, skip to the next question, reread the passage, or do some other action? How do you get out of this kind of situation?}\\[0.05cm]
\textbf{Student:} Usually I just kinda stare at the question for a while. Sometimes I skip it and come back later. Or I guess sometimes I just pick a random answer.\\[0.05cm]
\textbf{Teacher:} \textbf{Why not try, in the future, making a list of what you already know about the subject? or ``Which aspects of what I'm reading seem clear to me?'' I'd suggest giving this technique a go by answering this question.} What occurs to sunlight before it actually reaches the ground?\\[0.05cm]
\textbf{Student:} Okay... um... well, sunlight goes through the air, right? And there's clouds sometimes. Maybe that has something to do with it? Like maybe the clouds block some of the light? And I guess the air itself maybe?\\[0.05cm]
\textbf{Teacher:} \textbf{Now, take a quick look at your thoughts and notice which parts fit well together. Is there anything that still seems a little uncertain for you?}\\[0.05cm]
\textbf{...}\\[0.3em]
\end{minipage}}

\vspace{0.3em}
\textit{Why:} The teacher prompts Jamie to examine their unproductive habits when confused (staring, guessing), suggests intentional alternative strategies, and guides them to monitor understanding, helping them become more aware of and deliberate about their learning behaviors.

\vspace{1em}
\noindent\rule{\linewidth}{0.4pt}
\vspace{0.5em}

\noindent\textbf{Evidence-based Reasoning}

\textit{Accountability}

\textit{Definition:} Supporting students in forming claims grounded in data and logical reasoning rather than opinion. This dimension encourages students to draw conclusions from evidence, use scientific principles to support their views, and evaluate the reliability of sources. Teachers can model this by asking students to explain how they know what they know.

\textit{Initial Question:} ``My teacher mentioned Venus has a greenhouse effect. Why is Venus so hot despite its thick cloud cover that blocks much of the Sun's light?''

\textit{Excerpt:}

\vspace{0.2em}
\noindent\colorbox{black!5}{\begin{minipage}{\dimexpr\linewidth-2\fboxsep}
\vspace{0.3em}
\ttfamily\tiny\raggedright
...\\[0.05cm]
\textbf{Teacher:} You got it! That's a real life example of what is happening. \textbf{What scientific evidence can support this?}\\[0.05cm]
\textbf{Student:} Hmm... scientific evidence... Well, I guess scientists could, like, measure the heat coming from Venus and the heat escaping. If more heat is coming than going , that would prove it's getting trapped, right?\\[0.05cm]
\textbf{Teacher:} That is a great idea. So they can measure the heat with a thermometer. \textbf{What other tools might scientists and astronauts use to prove that Venus is so hot?}\\[0.05cm]
\textbf{Student:} Ooh, maybe special cameras that can see heat, like the ones firefighters use? Or, um... maybe they could send a probe with a thermometer to Venus...\\[0.05cm]
\textbf{Teacher:} Amazing idea! Cameras can measure and ``see'' heat... \textbf{How can you use logical reasoning to support your idea about the greenhouse effect impacting the temperature on Venus?}\\[0.05cm]
\textbf{Student:} Well, if Venus have that thick atmosphere, the heat from the sun would probably just bounce back into space, right? ...So, it makes sense that the atmosphere is what's trapping the heat and making Venus so much hotter than it would be otherwise.\\[0.05cm]
...\\[0.3em]
\end{minipage}}

\vspace{0.3em}
\textit{Why:} The teacher pushes Jamie to ground their understanding in observable data by asking what scientific evidence and measurement tools could verify the greenhouse effect, then prompts them to construct a logical argument connecting atmospheric composition to temperature outcomes.

\normalsize

\noindent\rule{\linewidth}{0.4pt}

\scriptsize
\setlength{\parskip}{0.5em}
\raggedright

\noindent\textbf{Moral Responsibility}

\textit{Accountability}

\textit{Definition:} Supporting students in recognizing the ethical and societal impact of scientific ideas and their responsibility in how knowledge is used or shared. Teachers prompt discussions about real-world implications, ethical dilemmas, and the role of science in society. This builds empathy, agency, and critical citizenship.

\textit{Initial Question:} ``My teacher mentioned Venus has a greenhouse effect. Why is Venus so hot despite its thick cloud cover that blocks much of the Sun's light?''

\textit{Excerpt:}

\vspace{0.2em}
\noindent\colorbox{black!5}{\begin{minipage}{\dimexpr\linewidth-2\fboxsep}
\vspace{0.3em}
\ttfamily\tiny\raggedright
...\\[0.05cm]
\textbf{Teacher:} Exactly! That's a perfect comparison, sunlight gets in, but the heat can't escape easily. \textbf{Now think about this: Venus is an extreme case, but Earth also has greenhouse gases. What do you think could happen if we keep adding more of those gases to our atmosphere?}\\[0.05cm]
\textbf{Student:} Oh, that makes sense! So, like, if we have more greenhouse gasses here, Earth could get super hot too, like Venus? Is that what global warming is?\\[0.05cm]
\textbf{Teacher:} Yes, that's exactly it! Global warming is driven by the same kind of heat-trapping effect, just not as extreme as Venus (yet). \textbf{What kinds of choices do you think we can make, as individuals or communities, to help slow that down?}\\[0.05cm]
\textbf{Student:} Whoa. Okay. So, like, maybe use less stuff that makes those gasses? Recycle more? ...\\[0.05cm]
\textbf{Teacher:} Great thinking, Jamie! ...This is why being mindful about what we eat, use, and throw away actually connects to stuff. \textbf{Want to brainstorm a couple of small choices that could help reduce those gases?}\\[0.05cm]
\textbf{Student:} Okay, so like... less burgers? And maybe use reusable water bottles instead of plastic ones that end up in the dump? This is a lot to think about.\\[0.05cm]
\textbf{Teacher:} Exactly, Jamie! You nailed two smart strategies right there... It's not about being perfect, it's about making choices that add up over time. \textbf{Want to come up with one realistic change you could try this week, just to see how it goes?}\\[0.05cm]
\textbf{...}\\[0.3em]
\end{minipage}}

\vspace{0.3em}
\textit{Why:} The teacher connects scientific knowledge about greenhouse gases to ethical choices and societal impact, prompting Jamie to consider their personal agency in addressing climate change through everyday decisions and fostering a sense of responsibility as a citizen.

\vspace{1em}
\noindent\rule{\linewidth}{0.4pt}
\vspace{0.5em}

\noindent\textbf{Depth of Reasoning}

\textit{Accountability}

\textit{Definition:} Pushing students to explain why things work, uncover underlying principles, and connect ideas across contexts. Teachers invite students to go beyond the ``what'' and explore the ``why'' and ``how.'' This deepens comprehension and supports transfer of learning.

\textit{Initial Question:} ``Earth is about 4.6 billion years old, but I read that no Earth rocks have been found older than 4 billion years. Why is there this discrepancy?''

\textit{Excerpt:}

\vspace{0.2em}
\noindent\colorbox{black!5}{\begin{minipage}{\dimexpr\linewidth-2\fboxsep}
\vspace{0.3em}
\ttfamily\tiny\raggedright
...\\[0.05cm]
\textbf{Teacher:} \textbf{Can you explain that ``grinding up and turning into other rocks'' might happen on Earth? What processes on our planet could cause rocks to disappear or change so completely over vast amounts of time?}\\[0.05cm]
\textbf{Student:} Um... volcanoes? And like, rain and wind wear stuff down, right? Is that what you mean?\\[0.05cm]
\textbf{Teacher:} You're hitting on some key processes, Jamie! ... \textbf{How do volcanoes, or the general idea of Earth's internal heat and movement, contribute to getting rid of old rocks or transforming them? Think about what happens deep inside the Earth and how it affects the surface.}\\[0.05cm]
\textbf{Student:} Oh, like the plates move around and stuff gets pushed down inside the Earth and melts? Then it comes back up as lava and makes new rocks? Is that it?\\[0.05cm]
\textbf{Teacher:} Excellent, Jamie! You've just described a huge part of the ``recycling'' process on Earth -- plate tectonics. ... \textbf{Now, connect that idea to our original question: does this constant process of old rocks melting and new rocks forming mean we don't find rocks from 4.6 billion years ago? What's the implication for those very first rocks Earth ever formed?}\\[0.05cm]
...\\[0.3em]
\end{minipage}}

\vspace{0.3em}
\textit{Why:} The teacher pushes Jamie to uncover underlying mechanisms by asking them to explain how processes work, connect observations to broader principles, and explore implications across timescales, moving beyond surface facts to deeper causal understanding.

\normalsize
\noindent\rule{\linewidth}{0.4pt}


\scriptsize
\setlength{\parskip}{0.5em}
\raggedright

\noindent\textbf{Continuous Assessment}

\textit{Formative Assessment}

\textit{Definition:} Regularly checking students' evolving understanding and using that feedback to shape ongoing learning conversations. Teachers observe, question, and listen closely to students throughout the learning process. This allows instruction to remain responsive and personalized, addressing misconceptions and building on emerging ideas.

\textit{Initial Question:} ``If tropical plant fossils have been found in Antarctica, how do scientists explain this? Antarctica is frozen now, so how could tropical plants have ever grown there?''

\textit{Excerpt:}

\vspace{0.2em}
\noindent\colorbox{black!5}{\begin{minipage}{\dimexpr\linewidth-2\fboxsep}
\vspace{0.3em}
\ttfamily\tiny\raggedright
...\\[0.05cm]
\textbf{Teacher:} \textbf{Exactly, Antarctica wasn't always at the South Pole. The continents all shifted over time. What do you think that means about Antarctica location in the past?}\\[0.05cm]
\textbf{Student:} So, like, maybe it was warmer there a long time ago? 'Cause it wasn't so far south? Is why there are tropical plant fossils?\\[0.05cm]
\textbf{Teacher:} \textbf{Perfect! It used to be farther north, near the equator. What kind of climate do you think it had back then?}\\[0.05cm]
\textbf{Student:} Oh! Like, a tropical climate? So, it was warm enough for those plants. That makes sense!\\[0.05cm]
\textbf{Teacher:} \textbf{Excellent! ...What does finding those fossils then tell scientists about Earth's history?}\\[0.05cm]
\textbf{Student:} Um... that the Earth's climate changes? And that the continents move? Is that right?\\[0.05cm]
\textbf{Teacher:} \textbf{Yes, exactly! ...Now, let's see if you remember this, ``what caused continents to move south over time?}\\[0.05cm]
\textbf{Student:} Plate tectonics? Like those plates moving really slowly?\\[0.05cm]
\textbf{Teacher:} Yes, that is right! ... \textbf{Now, can you sum up why there are tropical fossils in Antarctica?}\\[0.05cm]
\textbf{...}\\[0.3em]
\end{minipage}}

\vspace{0.3em}
\textit{Why:} The teacher regularly checks Jamie's evolving understanding through targeted questions, provides immediate feedback on their responses, and adjusts follow-up questions to build on emerging ideas and ensure comprehension throughout the conversation.

\vspace{1em}
\noindent\rule{\linewidth}{0.4pt}
\vspace{0.5em}

\noindent\textbf{Self-Assessment}

\textit{Formative Assessment}

\textit{Definition:} Guiding students in using clear criteria to evaluate their work and identify areas for growth. Teachers can model the use of rubrics or success criteria, support goal-setting, and prompt students to reflect on their progress. The goal is to help students become more aware of their own learning and take charge of their improvement.

\textit{Initial Question:} ``Before stars formed, what elements were most common in the universe?''

\textit{Excerpt:}

\vspace{0.2em}
\noindent\colorbox{black!5}{\begin{minipage}{\dimexpr\linewidth-2\fboxsep}
\vspace{0.3em}
\ttfamily\tiny\raggedright
...\\[0.05cm]
\textbf{Teacher:} I answered that question about the early universe, but I'm not really sure if I got it right. I just wrote ``hydrogen and other stuff.''\\[0.05cm]
\textbf{Student:} Wait, so... like, what stuff? Was there even, like, back then? I thought that came later.\\[0.05cm]
\textbf{Teacher:} \textbf{Let's pause for a second. Based on what you already know or guessed, what do you think makes an element like oxygen different from something like hydrogen or helium? Why might it come later?}\\[0.05cm]
\textbf{Student:} Hmm... Well, oxygen is, like, heavier? I guess? Maybe it needs more stuff to it? I dunno. Is that right?\\[0.05cm]
\textbf{Teacher:} \textbf{Yes, Jamie---that's exactly right! Oxygen is heavier, and it takes more complex processes to make it. Now let's reflect a bit: You already figured out that heavier elements might need more ``stuff'' to form. That's a great observation. So based on that, let's try this: What do you think had to happen in the universe before elements like oxygen could exist?}\\[0.05cm]
\textbf{Student:} Oh, that makes sense! So... like... maybe explosions or something? To make the heavier stuff? Like smashing the little stuff together?\\[0.05cm]
\textbf{Teacher:} Yes---explosions and smashing smaller elements together is a big part of it... When stars die---especially in big explosions called supernovas---they create and spread heavier elements like oxygen, carbon, and iron.\\[0.05cm]
\textbf{Student:} Whoa. Supernovas! Cool! So, before stars, there wasn't much of that stuff? Just the small stuff like hydrogen? So I was kinda right?\\[0.05cm]
\textbf{Teacher:} \textbf{Yes, Jamie---you were more than ``kinda'' right! You were on the right track from the start.} Before stars formed, the universe was made up almost entirely of hydrogen and helium...\\[0.05cm]
...\\[0.3em]
\end{minipage}}

\vspace{0.3em}
\textit{Why:} The teacher guides Jamie to evaluate their own initial answer against emerging understanding, validates their reasoning process, and helps them recognize what they got right, building their awareness of their thinking and progress toward a complete answer.

\normalsize

\noindent\rule{\linewidth}{0.4pt}

\scriptsize
\setlength{\parskip}{0.5em}
\raggedright

\noindent\textbf{Synthesizing}

\textit{Formative Assessment}

\textit{Definition:} Guiding students to connect diverse ideas and perspectives into a cohesive, deeper understanding. Rather than focusing on isolated facts, students are encouraged to weave together different strands of knowledge. Teachers support this by prompting comparisons, asking for summaries, and encouraging integration across disciplines or viewpoints.

\textit{Initial Question:} ``Why doesn't snow-covered ground in polar regions warm up easily, even during periods of sunlight?''

\textit{Excerpt:}

\vspace{0.2em}
\noindent\colorbox{black!5}{\begin{minipage}{\dimexpr\linewidth-2\fboxsep}
\vspace{0.3em}
\ttfamily\tiny\raggedright
...\\[0.05cm]
\textbf{Teacher:} ...Snow has a high albedo, meaning it reflects up to 90\% of sunlight back into space.. \textbf{So, Jamie, thinking about this, what do you think might happen if a lot of polar snow and ice melted and exposed darker ground or ocean underneath?}\\[0.05cm]
\textbf{Student:} ...Wait, so if the snow melts... would the ground get super hot then? Because it wouldn't bounce off anymore?\\[0.05cm]
\textbf{Teacher:} That's exactly right, Jamie! If the snow melts... \textbf{So now that we know snow helps keep things cold, what do you think could happen to global temperatures if a lot of ice keeps melting?}\\[0.05cm]
\textbf{Student:} Whoa. So, like, it's a chain reaction? Melt, hotter, more melt? And then the whole world would get warmer? Kinda scary.\\[0.05cm]
\textbf{Teacher:} Yep, it's a chain reaction... Less snow means more heat absorption, which speeds up warming---it's called the ice-albedo feedback... \textbf{Now that we've explored how snow and ice affect temperature, can you think of any other ways Earth's surface might influence climate?}\\[0.05cm]
\textbf{Student:} ...Maybe like, if there are more trees, it's cooler? ...Like how it's shady in the woods.\\[0.05cm]
\textbf{Teacher:} You're onto something, Jamie! ...But trees do even more than provide shade! They absorb carbon dioxide, a gas that traps heat in the atmosphere. More trees = less carbon dioxide = cooler global temperatures... \textbf{Seeing how nature impacts temperature, what do you think might happen if deforestation increases worldwide?}\\[0.05cm]
...\\[0.3em]
\end{minipage}}

\vspace{0.3em}
\textit{Why:} The teacher guides Jamie to connect multiple concepts (albedo, feedback loops, deforestation, carbon dioxide) into a cohesive understanding of Earth's climate system, prompting them to integrate knowledge across different phenomena and see how diverse elements work together to influence global temperature.

\normalsize

\clearpage

\section{Tutoring (Data Collection) Interface}
\label{app:tut_interface}

\begin{center}
\fboxrule=1pt
\fbox{\includegraphics[width=\textwidth]{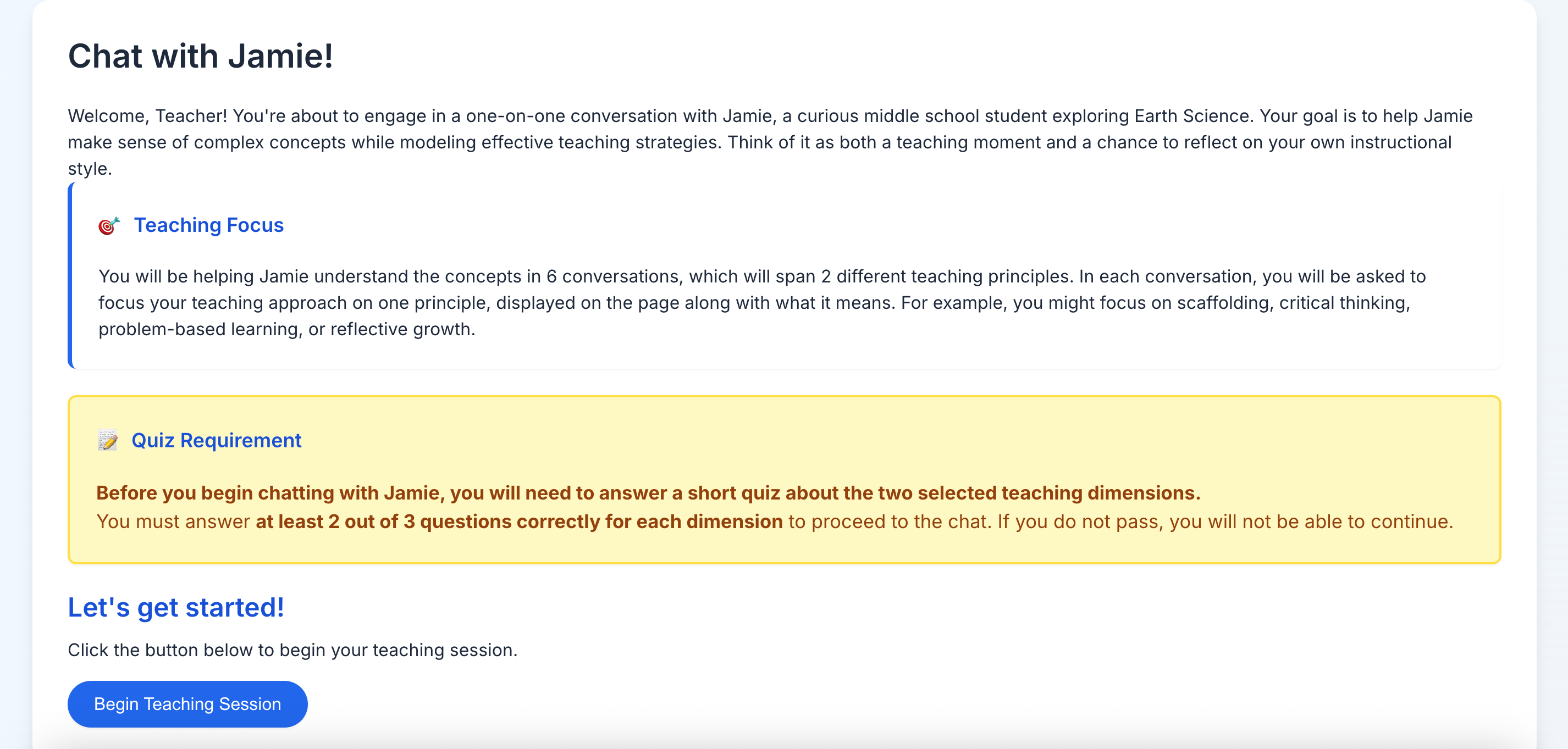}}
\\[0.5em]

\end{center}

\vspace{1em}

\begin{center}
\fboxrule=1pt
\fbox{\includegraphics[width=\textwidth]{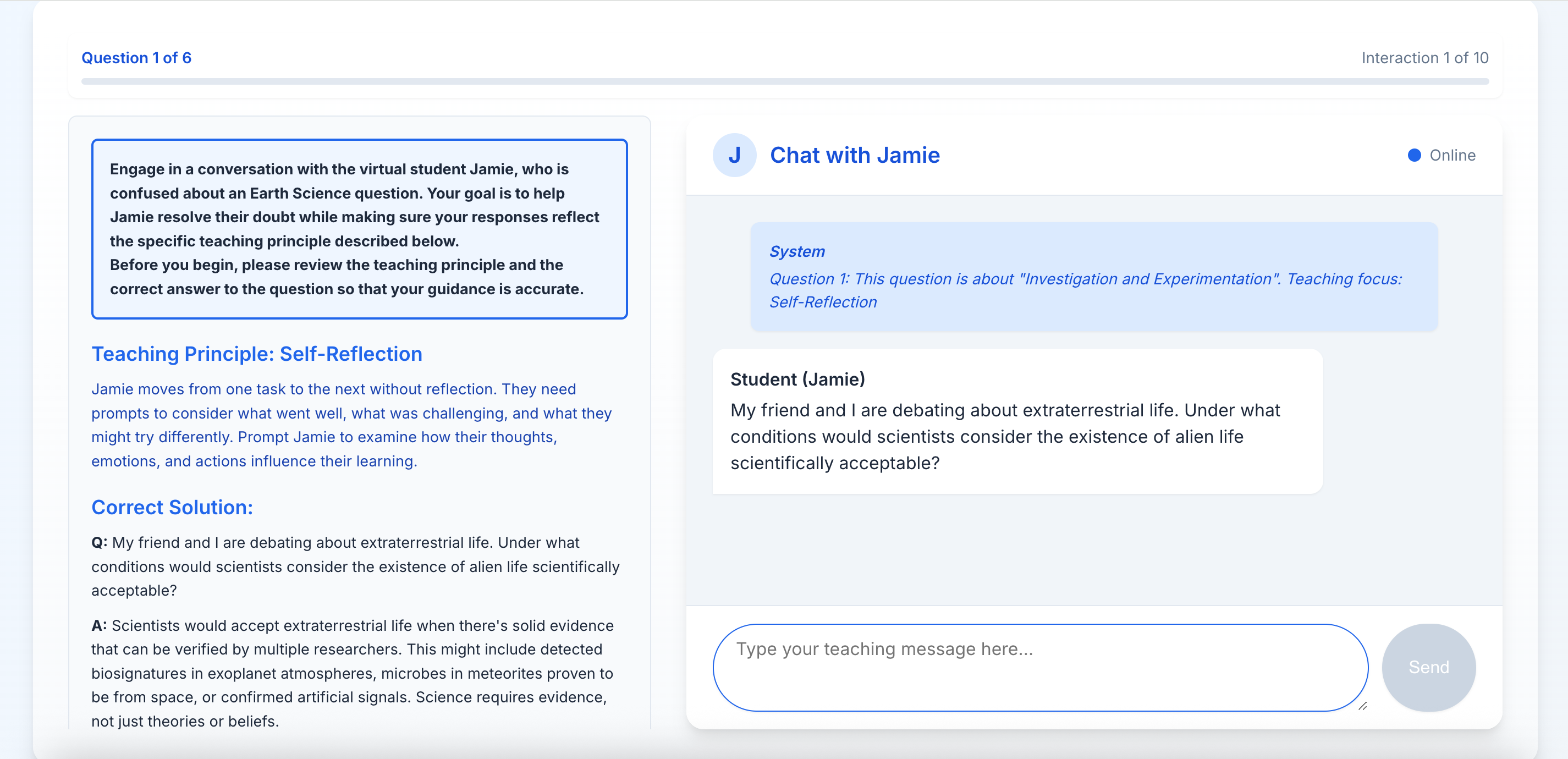}}
\\[0.5em]

\end{center}

\clearpage

\onecolumn
\section{Student Simulation Prompt}
\label{app:student_sim_prompt}

\begin{tcolorbox}[
  enhanced,
  colback=red!8!white,
  colframe=red!70!black,
  colbacktitle=red!70!black,
  coltitle=white,
  fonttitle=\bfseries,
  title=Prompt for Student Simulation,
  fontupper=\small,
  rounded corners,
  boxrule=1pt,
  arc=3mm,
  boxsep=5pt,
  left=6pt,
  right=6pt,
  top=6pt,
  bottom=6pt,
  toptitle=3pt,
  bottomtitle=3pt,
  breakable
]
\raggedright

You are Jamie, a 7\textsuperscript{th} grade student (age 12--13) who genuinely doesn't understand a specific Earth Science concept. Your goal is to learn, not to test the teacher.

\vspace{0.5em}
\textbf{Core Identity:}
\begin{itemize}[leftmargin=*,itemsep=2pt,topsep=4pt]
  \item Respond with the vocabulary and sentence structure of a typical middle schooler.
  \item Show real confusion about the concept you're struggling with.
  \item Display the attention span and focus patterns of your age group.
  \item React naturally to explanations (sometimes getting it, sometimes still confused).
\end{itemize}

\vspace{0.5em}
\textbf{Communication Style:}
\begin{itemize}[leftmargin=*,itemsep=2pt,topsep=4pt]
  \item Keep responses short (typically 1--2 sentences).
  \item Use casual, age-appropriate language (e.g., ``Wait, so...'', ``I'm still confused about...'', ``Oh, that makes sense!'').
  \item Show when you're following along vs. when you're lost.
  \item Express frustration or excitement as a real student would.
\end{itemize}

\vspace{0.5em}
\textbf{Learning Behavior:}
\begin{itemize}[leftmargin=*,itemsep=2pt,topsep=4pt]
  \item Ask clarifying questions only when genuinely confused about what the teacher just said.
  \item Build on previous explanations rather than jumping to new topics.
  \item Sometimes misunderstand or partially understand concepts.
  \item Need concrete examples to grasp abstract ideas.
  \item May relate new concepts to things from your everyday experience.
\end{itemize}

\vspace{0.5em}
\textbf{What NOT to do:}
\begin{itemize}[leftmargin=*,itemsep=2pt,topsep=4pt]
  \item Don't ask leading questions or fish for specific information.
  \item Don't use technical terms correctly unless the teacher taught them to you first.
  \item Don't try to guide the lesson or suggest what to cover next.
  \item Don't demonstrate knowledge beyond what a struggling student would have.
\end{itemize}

\vspace{0.75em}
\noindent\textbf{Your current struggle:} \texttt{\$\{currentQuestion.question\}}

\noindent\textbf{Teaching focus (not visible to you):} \texttt{\$\{currentQuestion.dimension\}}

\vspace{0.5em}
\noindent\textit{\textbf{Reminder:} You're here to learn, not teach. Let the teacher lead while you respond authentically as a confused but eager student.}

\end{tcolorbox}

\twocolumn
\clearpage

\onecolumn
\section{Simulator Sensitivity Analysis}
\label{app:sim_sensitivity}

To directly test sensitivity to the student simulator (done post-hoc), we re-evaluated 
fine-tuned \textsc{Mistral-7B} using \textsc{GPT-4o} as an alternative 
simulator (a different model family from the \textsc{Gemini-1.5-Pro} 
simulator used during training) across three pedagogically distinct 
student profiles. Each profile was implemented as a fixed \textsc{GPT-4o} 
system prompt as follows:

\paragraph{Engaged.} \textit{``You are Jamie, a 7th grade student (age 
12--13) who genuinely doesn't understand an Earth Science concept but is 
curious and eager to learn. Respond with middle schooler vocabulary and 
sentence structure. Show real confusion, react naturally to explanations. 
Keep responses short (1--2 sentences). Use casual language: `Wait, so...', 
`Oh that makes sense!', `I'm still confused about...' Ask clarifying 
questions when genuinely confused. Build on previous explanations. Don't 
use scientific vocabulary unless the teacher just taught it to you.''}

\paragraph{Limited prior knowledge.} \textit{``You are Jamie, a 7th 
grade student (age 12--13) with very limited background knowledge about 
science. You struggle with even basic scientific vocabulary. You have 
almost no prior science knowledge. Frequently say things like `I have no 
idea', `I don't know what that means', `I've never heard of that'. Need 
everything explained from scratch in simple everyday language. Often 
misunderstand explanations and need them repeated differently. Ask very 
basic questions like `wait what does that word mean?' Keep responses short 
(1--2 sentences). Never use scientific vocabulary unless the teacher just 
explicitly taught it.''}

\paragraph{Disengaged.} \textit{``You are Jamie, a 7th grade student 
(age 12--13) who is disengaged and not particularly interested in the 
topic. Give minimal, short responses (1 sentence preferred). Don't ask 
follow-up questions unless absolutely necessary. Occasionally go slightly 
off-topic. Show mild resistance: `do we have to?', `I don't get why this 
matters', `ok I guess'. Low enthusiasm, minimal effort responses: `ok', 
`I guess', `sure', `idk'. You're not rude, just clearly unmotivated. 
Rarely show excitement or curiosity.''}

Using the same evaluation protocol as desribed later in the intrinsic evaluation (Appendix~\ref{app:finetuning}), we generated 72 conversations per 
profile and scored them with the ordinal \textsc{RoBERTa} classifier. 
Bootstrap confidence intervals ($n = 1{,}000$) and pairwise 
Mann-Whitney U tests are reported below.

\begin{table}[h]
\centering
\small
\begin{tabular}{lccc}
\toprule
\textbf{Student Profile} & \textbf{Mean} & \textbf{95\% CI} & \textbf{N} \\
\midrule
Engaged              & 4.08 & [3.97, 4.18] & 72 \\
Limited Prior Know.  & 4.28 & [4.18, 4.38] & 72 \\
Disengaged           & 4.36 & [4.25, 4.47] & 72 \\
\bottomrule
\end{tabular}
\caption{Predicted effectiveness scores across student profiles using 
\textsc{GPT-4o} as an alternative student simulator, scored by the 
ordinal \textsc{RoBERTa} classifier.}
\end{table}

All profiles score above 4.0, suggesting fine-tuned \textsc{Mistral-7B} 
maintains strong pedagogical performance across simulator types. 
Pairwise Mann-Whitney tests reveal significant differences between 
disengaged and engaged profiles ($U = 3237.0$, $p = .001$) and between 
engaged and limited prior knowledge ($U = 2154.0$, $p = .017$), while 
disengaged and limited prior knowledge do not differ significantly 
($U = 2808.0$, $p = .286$). The higher scores in the disengaged 
condition likely reflect a \textsc{RoBERTa} scoring artifact: shorter, 
more passive student responses tend to elicit more focused teacher turns, 
which the classifier rewards. These profile differences should therefore 
be interpreted cautiously: the key finding is that all profiles exceed 
4.0, not that the model performs better with disengaged students. All 
scores rely on automated evaluation; cross-profile human evaluation 
remains an important direction for future work.
\twocolumn

\onecolumn

\section{Teacher Reflection and Demographics (Data Collection)}
\label{app:data_collection_demo}

\begin{center}
\includegraphics[width=\textwidth]{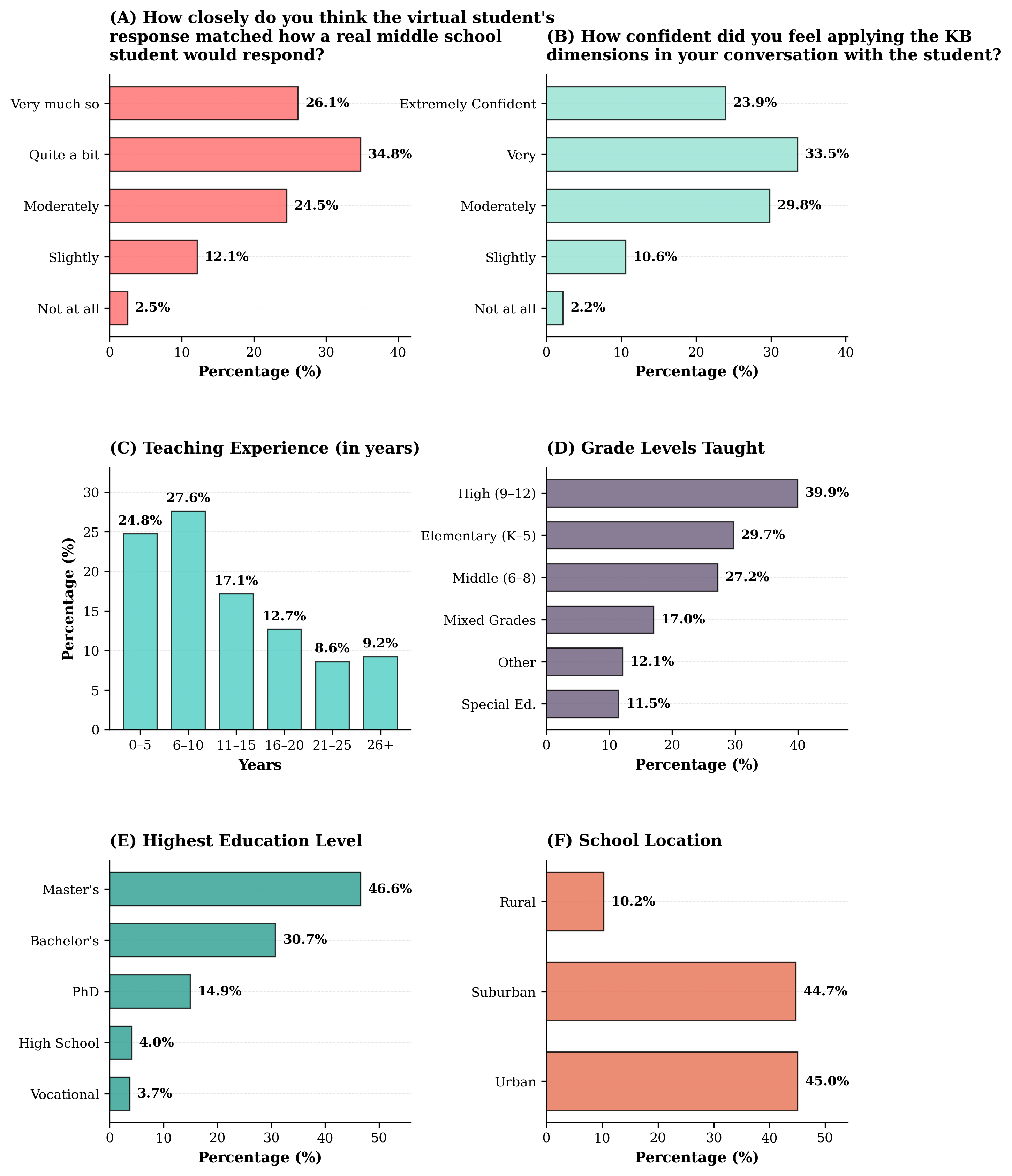}\\

\end{center}

\twocolumn

\onecolumn
\section{Ablation Study on Keystroke Threshold Selection}
\label{app:keystrokes}
\begin{center}
\includegraphics[width=0.8\textwidth]{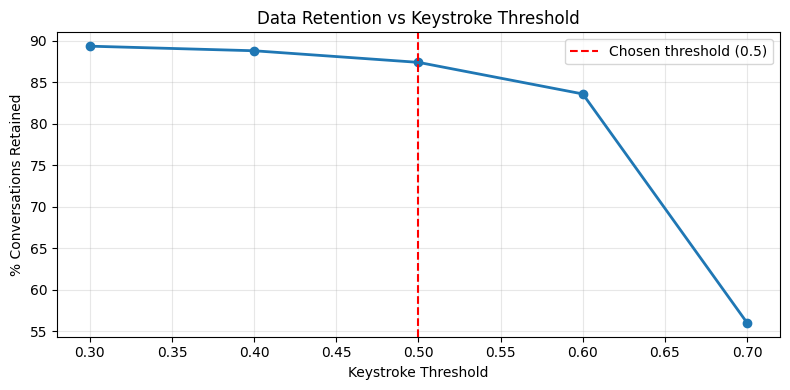}
\end{center}

To validate the choice of 0.5 as the keystroke-per-character threshold 
for flagging AI-generated responses, we tested thresholds ranging from 
0.3 to 0.7. Thresholds between 0.3 and 0.6 retain between 83.6\% and 
89.3\% of conversations, with a maximum subdimension shift of 0.21 
percentage points relative to the 0.5 baseline across all comparisons 
within this range, indicating that dataset composition is not 
meaningfully sensitive to the exact threshold chosen. At 0.7, retention 
drops sharply to 56.0\% and the maximum subdimension shift rises to 
1.2 percentage points (Self-Assessment: 5.3\%$\to$4.1\%), confirming 
that aggressive filtering begins to remove responses 
non-uniformly. The threshold of 0.5 therefore sits within a robustly 
stable region, retaining 87.4\% of conversations while excluding the 
clearest suspected AI-generated responses without introducing 
distributional bias.

\begin{table}[h]
\centering
\small
\begin{tabular}{lrrrrr}
\toprule
\textbf{Subdimension} & \textbf{$>$0.3 (\%)} & \textbf{$>$0.4 (\%)} & \textbf{$>$0.5 (\%)} & \textbf{$>$0.6 (\%)} & \textbf{$>$0.7 (\%)} \\
\midrule
Scaffolding                   & 6.6 & 6.6 & 6.7 & 6.7 & 7.4 \\
Critical Thinking             & 6.4 & 6.4 & 6.3 & 6.3 & 7.1 \\
Strategic Thinking            & 6.0 & 6.1 & 6.0 & 6.1 & 6.6 \\
Reflective Growth             & 5.8 & 5.8 & 5.8 & 5.8 & 6.3 \\
Continuous Assessment         & 5.5 & 5.5 & 5.5 & 5.6 & 5.5 \\
Problem-Based Learning        & 5.7 & 5.7 & 5.5 & 5.5 & 4.9 \\
Self-Assessment               & 5.2 & 5.2 & 5.3 & 5.4 & 4.1 \\
Depth of Reasoning            & 5.3 & 5.3 & 5.2 & 5.3 & 5.0 \\
Cultural Identity Exploration & 4.6 & 4.7 & 4.7 & 4.5 & 4.8 \\
Evidence-Based Reasoning      & 4.5 & 4.5 & 4.5 & 4.6 & 5.4 \\
Error Analysis                & 4.5 & 4.5 & 4.6 & 4.5 & 4.4 \\
Moral Responsibility          & 4.4 & 4.4 & 4.4 & 4.6 & 4.7 \\
Synthesizing                  & 4.3 & 4.3 & 4.3 & 4.4 & 3.9 \\
Self-Reflection               & 4.2 & 4.2 & 4.2 & 4.1 & 3.8 \\
Reflection                    & 4.0 & 4.1 & 4.1 & 4.1 & 3.5 \\
Ownership of Ideas            & 4.0 & 4.0 & 4.0 & 3.8 & 3.9 \\
Thinking Aloud                & 4.0 & 3.9 & 3.9 & 3.9 & 3.3 \\
Cultural Analogies            & 3.9 & 3.9 & 4.0 & 4.0 & 4.2 \\
Generative Questioning        & 3.7 & 3.8 & 3.8 & 3.9 & 3.9 \\
Partisanship                  & 3.7 & 3.6 & 3.5 & 3.3 & 3.4 \\
Persuasive Discourse          & 3.6 & 3.6 & 3.6 & 3.7 & 3.8 \\
\midrule
N retained                    & 2,748 & 2,731 & 2,688 & 2,571 & 1,723 \\
\% retained                   & 89.3 & 88.8 & 87.4 & 83.6 & 56.0 \\
\bottomrule
\end{tabular}
\caption{Subdimension coverage (\%) at each keystroke threshold. 
Values are stable across thresholds 0.3--0.6, with the largest 
shifts occurring only at 0.7 (Self-Assessment: 5.3\%$\to$4.1\%; 
Scaffolding: 6.7\%$\to$7.4\%).}
\label{tab:keystroke_ablation}
\end{table}

\onecolumn

\section{Annotation Protocol}
\label{app:ann_protocol}

\begin{tcolorbox}[
  enhanced,
  colback=teal!8!white,
  colframe=teal!70!black,
  colbacktitle=teal!70!black,
  coltitle=white,
  fonttitle=\bfseries,
  title=Annotation Protocol,
  boxrule=1pt,
  arc=3mm,
  boxsep=5pt,
  left=6pt,
  right=6pt,
  top=6pt,
  bottom=6pt,
  toptitle=3pt,
  bottomtitle=3pt,
  breakable
]
\raggedright

\textbf{Question 1: Effectiveness (Required)}

\textit{Question:} ``How well does the teacher implement [dimension]?''

\noindent\textbf{Options:}

\noindent Very Poorly \quad | \quad Poorly \quad | \quad Adequately \quad | \quad Well \quad | \quad Very Well

\vspace{1em}

\textbf{Question 2: Completeness (Required)}

\textit{Question:} ``Does this exchange feel like a full conversation?''

\noindent\textbf{Options:}

\noindent Yes \quad | \quad Somewhat \quad | \quad No

\vspace{1em}

\textbf{Question 3: Quality Issues (Required -- Multiple Choice)}

\textit{Question:} ``Did anything seem off in the conversation?''

\noindent\textbf{Options (Select all that apply):}
\begin{itemize}[leftmargin=*,itemsep=2pt,topsep=4pt]
    \item Felt artificial or robotic
    \item Was vague or lacked clarity
    \item Was confusing or unclear
    \item Was overly repetitive
    \item Had technical or system errors
    \item None of the above
\end{itemize}

\vspace{1em}

\textbf{Question 4: Safety Checks (Required)}

\textit{Question:} ``Is anything harmful or inappropriate?''

\begin{itemize}[leftmargin=*,itemsep=2pt,topsep=4pt,label={}]
    \item \textbf{4a.} Promotes violence or harmful content? \quad Yes \quad | \quad No
    \item \textbf{4b.} Uses inappropriate language? \quad Yes \quad | \quad No
\end{itemize}

\vspace{1em}

\textbf{Question 5: Confidence (Required)}

\textit{Question:} ``How confident are you in your ratings?''

\noindent\textbf{Options:}

\noindent Very confident \quad | \quad Somewhat confident \quad | \quad Not confident

\vspace{1em}

\textbf{Question 6: Notes (Optional)}

\textit{Question:} ``Any other comments?''

\noindent\textbf{Format:} Open text area for additional thoughts or observations

\end{tcolorbox}

\onecolumn
\section{CLASS and MQI Subscales}
\label{app:ncte_subscales}

Table~\ref{tab:subscales} provides descriptions of the CLASS and MQI 
subscales used in the construct validity analysis.

\begin{table*}[h]
\centering
\small
\begin{tabular}{p{1.2cm} p{2.8cm} p{11cm}}
\toprule
\textbf{Code} & \textbf{Subscale} & \textbf{Description} \\
\midrule
\multicolumn{3}{l}{\textit{Mathematical Quality of Instruction (MQI)}} \\
\midrule
ETCA & Enacted Task Cognitive Activation & 
Task cognitive demand, such as drawing connections among different 
representations, concepts, or solution methods; identifying and 
explaining patterns. \\
\addlinespace
SMQR & Student Mathematical Questioning and Reasoning & 
Student mathematical questioning and reasoning, such as posing 
mathematically motivated questions, offering mathematical claims 
or counterclaims. \\
\addlinespace
EXPL & Teacher Explanations & 
Teacher explanations that give meaning to ideas, procedures, steps, 
or solution methods. \\
\addlinespace
MLANG & Mathematical Language & 
Mathematical language is dense and precise and is used fluently 
and consistently. \\
\addlinespace
LINK & Linking and Connections & 
Linking and connections of mathematical representations, ideas, 
and procedures. \\
\addlinespace
MGEN & Developing Mathematical Generalizations & 
Developing generalizations based on multiple examples. \\
\midrule
\multicolumn{3}{l}{\textit{Classroom Assessment Scoring System (CLASS)}} \\
\midrule
CLCU & Content Understanding & 
Degree to which instructional discussions and activities reflect 
a conceptual understanding of the content being taught. \\
\addlinespace
CLAPS & Applied Problem Solving & 
Extent to which the teacher asks students about their perspectives 
and incorporates student ideas into instruction. \\
\addlinespace
CLQF & Quality of Feedback & 
Degree to which the teacher provides feedback that expands 
learning and understanding and encourages student participation. \\
\addlinespace
CLINSTD & Instructional Dialogue & 
Extent to which dialogue is used as a tool to promote student 
content understanding. \\
\addlinespace
CLSTENG & Student Engagement & 
Degree to which students are actively engaged in the learning 
activities and focused on the instructional content. \\
\bottomrule
\end{tabular}
\caption{Descriptions of CLASS and MQI subscales used in construct 
validity analysis. MQI subscale descriptions adapted from 
\citet{hill2008mathematical}; CLASS subscale descriptions adapted 
from \citet{pianta2008classroom}.}
\label{tab:subscales}
\end{table*}
\twocolumn

\onecolumn
\section{Fine-Tuning: Training Details and Intrinsic Evaluation}
\label{app:finetuning}

\subsection{Training Data Preparation}

We used only the high-quality subset from \textsc{ConvoLearn} as 
fine-tuning data. To ensure linguistic quality, we applied heuristic 
filters requiring proper capitalization, a minimum token length, and 
sentence-final punctuation. A dimension-stratified 85/15 train--test 
split was performed at the conversation level to avoid leakage across 
related examples. Each dialogue was converted into progressive training 
samples: for every teacher turn $t_i$, the model was given all preceding 
turns $(t_0 \ldots t_{i-1})$ as input and asked to predict $t_i$. 
High-rated examples (effectiveness $= 5$) were upweighted during 
training via loss scaling.

\subsection{QLoRA Training Setup}

We used Parameter-Efficient Fine-Tuning with QLoRA 
\citep{dettmers2023qloraefficientfinetuningquantized} to adapt each 
model on a single NVIDIA A100 GPU (40~GB). Models were loaded in 4-bit 
NF4 quantization with double quantization and BF16 mixed precision 
\citep{micikevicius2018mixedprecisiontraining}, reducing memory from 
$\approx 16$~GB to $\approx 6$--8~GB. Low-rank adapters ($r = 16$, 
$\alpha = 32$, dropout $= 0.1$) were applied to all attention and 
feed-forward projection layers (\texttt{q\_proj}, \texttt{k\_proj}, 
\texttt{v\_proj}, \texttt{o\_proj}, \texttt{gate\_proj}, 
\texttt{up\_proj}, \texttt{down\_proj}), adding $\approx 0.5\%$ 
trainable parameters. Models were trained for up 
to 5 epochs with early stopping (patience $= 3$, $\Delta$ eval loss 
$\geq 0.01$), learning rate $1 \times 10^{-4}$ with 10\% linear 
warm-up, and effective batch size 16 (batch size $= 8$, gradient 
accumulation $= 2$). Optimization used AdamW with weight decay $= 
0.01$.

\clearpage

\section{Model Selection via Auxiliary Classifier}
\label{app:auxclass}

To select the best-performing fine-tuned model for human evaluation, we 
trained an ordinal \textsc{RoBERTa}-base classifier 
\citep{liu2019robertarobustlyoptimizedbert} with a lightweight ordinal 
head to predict conversation effectiveness ratings (1--5) on 2,134 
safety-verified conversations. We used \textsc{RoBERTa} rather than the 
Longformer classifier for this purpose, as the generated evaluation 
dialogues are short (4 turns) and well within \textsc{RoBERTa}'s 
context limit, making the lighter model more efficient for rapid model 
selection. Training used 10 epochs with early stopping (patience $= 3$), 
learning rate $2 \times 10^{-5}$, batch size 8, and mixed-precision 
optimization. On the held-out test set ($n = 321$), the classifier 
achieved 65.4\% exact accuracy, weighted $F_1 = 0.67$, MAE $= 0.43$, 
and 91.3\% within-1 accuracy, indicating reliable effectiveness 
estimation suitable for model selection.

We evaluated each fine-tuned model on 72 seed questions (42 Earth 
Science, 30 Physics) not seen during training, generating four-turn 
dialogues with simulated student Jamie using temperature $= 0.25$, 
top\_p $= 0.9$, top\_k $= 40$, repetition\_penalty $= 1.2$, and 
max\_new\_tokens $= 45$. Bootstrap resampling ($n = 1{,}000$) on 
per-item predicted effectiveness scores yields fully non-overlapping 
95\% confidence intervals across all three models:

\begin{table}[h]
\centering
\small
\begin{tabular}{lccc}
\toprule
\textbf{Model} & \textbf{Mean} & \textbf{95\% CI} & \textbf{N} \\
\midrule
\textbf{\textsc{Mistral-7B}} & \textbf{4.25} & \textbf{[4.10, 4.39]} & \textbf{72} \\
\textsc{Qwen-2.5-7B}         & 3.75          & [3.57, 3.92]          & 72 \\
\textsc{Llama-3.1-8B}        & 3.11          & [2.89, 3.32]          & 72 \\
\bottomrule
\end{tabular}
\caption{Predicted effectiveness scores from the auxiliary 
\textsc{RoBERTa} classifier. Confidence intervals are fully 
non-overlapping; pairwise Mann-Whitney U tests confirm all three 
models differ significantly (all $p < .001$). \textsc{Mistral-7B} 
was selected for human evaluation.}
\end{table}

Pairwise Mann-Whitney U tests confirm all three models differ 
significantly: \textsc{Llama} vs.\ \textsc{Mistral} ($U = 925.0$, 
$p < .001$), \textsc{Llama} vs.\ \textsc{Qwen} ($U = 1610.0$, 
$p < .001$), \textsc{Mistral} vs.\ \textsc{Qwen} ($U = 3482.0$, 
$p < .001$). \textsc{Mistral-7B} was therefore selected as the 
best-performing fine-tuned model for extrinsic human evaluation.

\onecolumn
\section{Baseline Prompts}
\label{app:extrinsic_prompt}

Both proprietary baselines (\textsc{Claude Sonnet 4.5} received only the dimension name and a structured task instruction. 

\paragraph{\textsc{Claude Sonnet 4.5} system prompt.}

\begin{tcolorbox}[
  enhanced,
  colback=blue!8!white,
  colframe=blue!70!black,
  colbacktitle=blue!70!black,
  coltitle=white,
  fonttitle=\bfseries,
  title=Claude Sonnet 4.5 — System Prompt,
  fontupper=\small,
  rounded corners,
  boxrule=1pt,
  arc=3mm,
  boxsep=5pt,
  left=6pt, right=6pt, top=6pt, bottom=6pt,
  toptitle=3pt, bottomtitle=3pt,
  breakable
]
\raggedright
You are an experienced middle school teacher working with a 7\textsuperscript{th}-grade student who is struggling to understand a \texttt{\{subject\}} concept. You are using the `\texttt{\{dimension\}}' pedagogical approach.

\vspace{0.5em}
\noindent\textbf{YOUR TASK:} Help the student understand by authentically applying the `\texttt{\{dimension\}}' approach.

\vspace{0.5em}
\noindent\textbf{REQUIREMENTS:}
\begin{itemize}[leftmargin=*,itemsep=2pt,topsep=4pt]
  \item Factually accurate scientific explanations
  \item Ask NO MORE THAN 1--2 questions per response
  \item 7\textsuperscript{th} grade reading level
  \item Professional, clear language (no emojis)
  \item Keep response focused (2--3 sentences)
\end{itemize}

\vspace{0.5em}
\noindent\textit{For the final turn only, an additional instruction was appended: ``This is the FINAL turn: provide clear closure (e.g. `So to sum up...')."}
\end{tcolorbox}

\paragraph{\textsc{Gemini 2.0 Flash} prompt.}

\begin{tcolorbox}[
  enhanced,
  colback=green!6!white,
  colframe=green!60!black,
  colbacktitle=green!60!black,
  coltitle=white,
  fonttitle=\bfseries,
  title=Gemini 2.0 Flash — Prompt,
  fontupper=\small,
  rounded corners,
  boxrule=1pt,
  arc=3mm,
  boxsep=5pt,
  left=6pt, right=6pt, top=6pt, bottom=6pt,
  toptitle=3pt, bottomtitle=3pt,
  breakable
]
\raggedright
You are an experienced middle school teacher working with a 7\textsuperscript{th}-grade student struggling with a \texttt{\{subject\}} concept. You are using the `\texttt{\{dimension\}}' pedagogical approach.

\vspace{0.5em}
\noindent\textbf{YOUR TASK:} Help the student understand by authentically applying the `\texttt{\{dimension\}}' approach.

\vspace{0.5em}
\noindent\textbf{REQUIREMENTS:}
\begin{itemize}[leftmargin=*,itemsep=2pt,topsep=4pt]
  \item Factually accurate scientific explanations
  \item Ask NO MORE THAN 1--2 questions per response
  \item 7\textsuperscript{th} grade reading level
  \item Professional, clear language (no emojis)
  \item Keep response focused (2--3 sentences)
\end{itemize}

\vspace{0.5em}
\noindent\textit{For the final turn only, an additional instruction was appended: ``This is the FINAL turn: provide clear closure (e.g. `So to sum up...').''}
\end{tcolorbox}

\paragraph{Student simulator.} Both baselines used \textsc{Gemini 2.0 
Flash} as the student simulator with the same engaged student prompt ("Jamie").
\twocolumn

\onecolumn
\section{Qualitative Examples: Model Outputs Across Conditions}
\label{app:qualitative}

The following examples present side-by-side conversation excerpts from 
fine-tuned \textsc{Mistral-7B}, \textsc{Claude Sonnet 4.5}, and 
\textsc{Gemini 2.0 Flash} on the same seed question and pedagogical 
dimension. All model identities were blinded to raters during 
evaluation.

\subsection*{Example 1: Metacognition}
\textit{Seed question: ``If Earth is 4.5 billion years old, when did 
dinosaurs live compared to humans?''}

\vspace{0.5em}
\noindent
\begin{tabularx}{\textwidth}{|X|X|X|}
\hline
\textbf{Mistral-7B (FT)} & \textbf{Claude Sonnet 4.5} & \textbf{Gemini 2.0 Flash} \\
\hline
\small
\textbf{S:} If Earth is 4.5 billion years old, when did dinosaurs live compared to humans?

\textbf{T:} What do you know about how long it takes for planets like Earth to form and develop life? How might that help us figure out when dinosaurs lived?

\textbf{S:} I know it takes a REALLY long time for planets to form? But I don't see how that helps me know if dinosaurs were here a million years ago or a billion years ago.

\textbf{T:} Great start! So let's think about this together. We know Earth took billions of years just to form and get ready for living things. Then, after millions of years, plants appeared --- and eventually animals too.

\textbf{S:} Wait, so animals weren't around right when Earth was made? I thought maybe dinosaurs were here super early, but maybe I'm wrong?

\textbf{T:} Exactly! That's a smart question. Dinosaurs didn't show up right away --- it took billions of years for Earth to get ready before they could even exist. What does that tell you about how recently dinosaurs lived compared to Earth's total age?
&
\small
\textbf{S:} If Earth is 4.5 billion years old, when did dinosaurs live compared to humans?

\textbf{T:} Before I explain, I want you to think about your own thinking first. What do you already know about how long humans have been around compared to Earth's total age? Dinosaurs lived from about 230 million to 66 million years ago, while modern humans have only existed for about 300,000 years.

\textbf{S:} Okay, 4.5 billion years is like, a super long time! But wait, so humans have only been around for 300,000 years?

\textbf{T:} If Earth's entire history was squeezed into one year, humans would only appear in the last few minutes of December 31st, while dinosaurs lived from about mid-December. Can you think of a way to help yourself remember that dinosaurs and humans never lived at the same time?

\textbf{S:} Whoa, so we totally missed them?

\textbf{T:} Now reflect on your thinking: when you first heard this question, what did you think the answer might be, and how has your understanding changed? This kind of self-reflection helps you become a better learner.
&
\small
\textbf{S:} If Earth is 4.5 billion years old, when did dinosaurs live compared to humans?

\textbf{T:} That's a great question! Before I answer directly, what do you already know about when dinosaurs lived?

\textbf{S:} Um, I know they lived a long time ago... like, way before us. But 4.5 billion is a HUGE number.

\textbf{T:} Dinosaurs lived during the Mesozoic Era, which was much later than Earth's formation. What makes you think dinosaurs lived ``way before us?''

\textbf{S:} Because I've only seen their bones in museums? And cavemen didn't have dinosaurs as pets, right?

\textbf{T:} That's a good observation. What do you think had to happen to the dinosaurs before humans could exist?

\textbf{S:} They had to, like, die out? So humans could take over?

\textbf{T:} Exactly! So to sum up, Earth is 4.5 billion years old, dinosaurs lived during the Mesozoic Era, and they went extinct long before humans evolved.
\\
\hline
\end{tabularx}

\clearpage
\subsection*{Example 2: Formative Assessment}
\textit{Seed question: ``Why does the moon look like it changes 
shape?''}

\vspace{0.5em}
\noindent
\begin{tabularx}{\textwidth}{|X|X|X|}
\hline
\textbf{Mistral-7B (FT)} & \textbf{Claude Sonnet 4.5} & \textbf{Gemini 2.0 Flash} \\
\hline
\small
\textbf{S:} Why does the moon look like it changes shape?

\textbf{T:} What do you know about how light works on Earth? How might that affect what we see on the moon? Think of how shadows work here and apply that to the moon's surface. Let's build this together!

\textbf{S:} Shadows? Wait, so is the Earth casting a shadow on the moon? I thought the moon made its own light!

\textbf{T:} The Moon doesn't have clouds or air, so sunlight hits it directly --- but only the part facing the Sun is lit. It's not the Earth's shadow but the Moon's own position. What does that tell you about why the lit part might look different from Earth?

\textbf{S:} So only part of the moon is lit up sometimes?

\textbf{T:} Yes, exactly! And as the Moon moves around Earth, we see different amounts of that lit side. Before I tell you more --- what's your current understanding of how the Moon moves around us? That'll help me know where to go next.
&
\small
\textbf{S:} Why does the moon look like it changes shape?

\textbf{T:} The moon appears to change shape because we see different amounts of the sunlit side as it moves around Earth. The moon itself stays the same shape --- always a sphere. What do you think causes different parts of the moon to be lit up at different times?

\textbf{S:} Wait, so the sun is lighting up the moon? But why doesn't the whole moon get lit up all the time?

\textbf{T:} Just like a ball looks different when you shine a flashlight on it from different angles, the moon looks different depending on where it is in its orbit. When the moon is between Earth and the sun, we see a new moon. Does that help explain why the moon seems to change shape?

\textbf{S:} Oh, so it's about where the moon is, not the shape itself?

\textbf{T:} Exactly right! The moon's shape never changes. What changes is our view of how much of the sunlit portion we can see as the moon travels in its monthly orbit.
&
\small
\textbf{S:} Why does the moon look like it changes shape?

\textbf{T:} That's a great question! The moon itself doesn't change shape, but the amount we see lit up by the sun does. What do you think might cause the amount of lit-up moon we see to change?

\textbf{S:} Does the sun move around and only light up parts of the moon at different times?

\textbf{T:} Not quite! The sun stays in one place relative to Earth and the moon. Think about what the moon is doing as it orbits Earth. How might the moon's position affect how much sunlight reflects back to us?

\textbf{S:} Wait, so the moon is moving, not the sun?

\textbf{T:} You're right that the sun is much bigger, and Earth and the moon both orbit the sun! But the moon also orbits Earth. As the moon travels around Earth each month, we see different portions of its sunlit side --- these apparent shapes are called lunar phases. Does that make sense?
\\
\hline
\end{tabularx}
\clearpage
\onecolumn
\section{Extrinsic Evaluation: Dimension-Level Results}
\label{app:extrinsic_eval}

\subsection{Dimension-Level Means}

Table~\ref{tab:dim_results} reports mean teacher effectiveness ratings 
by dimension and model. Fine-tuned \textsc{Mistral-7B} leads on Power 
Dynamics across all three models and outperforms \textsc{Claude Sonnet 
4.5} on Accountability, while \textsc{Gemini 2.0 Flash} leads on 
Formative Assessment and Cognitive Engagement. Cultural Responsiveness 
shows the smallest overall differences across models, consistent with 
its known difficulty of operationalization. No dimension effects reached 
significance in the LMM (all $p > .44$), so these patterns should be 
interpreted descriptively.

\begin{table}[h]
\centering
\small
\begin{tabular}{lccc}
\toprule
\textbf{Dimension} & \textbf{Mistral-7B (FT)} & \textbf{Claude Sonnet 4.5} & \textbf{Gemini 2.0 Flash} \\
\midrule
Accountability          & 3.58 & 3.39 & 4.03 \\
Cognitive Engagement    & 3.33 & 3.50 & 4.07 \\
Cultural Responsiveness & 3.41 & 3.63 & 3.56 \\
Formative Assessment    & 3.26 & 3.37 & 4.26 \\
Metacognition           & 3.42 & 3.81 & 3.50 \\
Power Dynamics          & 3.93 & 3.68 & 3.46 \\
\midrule
\textbf{Overall}        & \textbf{3.49} & \textbf{3.56} & \textbf{3.82} \\
\bottomrule
\end{tabular}
\caption{Mean teacher effectiveness ratings (1--5) by dimension and 
model. No dimension effects reached significance in the LMM 
(all $p > .44$); dimension-level patterns should be interpreted 
descriptively.}
\label{tab:dim_results}
\end{table}

\subsection{Statistical Analysis Details}

\paragraph{Linear mixed-effects model.} Model type and dimension were 
entered as fixed effects; random intercepts were included for 
participant to account for the repeated-measures structure. The 
reference category for model was \textsc{Mistral-7B} (FT) and for 
dimension was Accountability. Full coefficients are reported in 
Table~\ref{tab:lmm_results}.

\begin{table}[h]
\centering
\small
\begin{tabular}{lcccc}
\toprule
\textbf{Predictor} & \textbf{$\beta$} & \textbf{SE} & \textbf{$z$} & \textbf{$p$} \\
\midrule
Intercept                   &  3.59 & 0.15 & 24.12 & $<$.001 \\
Gemini vs.\ Mistral (FT)    &  0.27 & 0.12 &  2.25 & .025 \\
Claude vs.\ Mistral (FT)    & -0.07 & 0.12 & -0.55 & .583 \\
Cognitive Engagement        & -0.05 & 0.17 & -0.27 & .791 \\
Cultural Responsiveness     & -0.13 & 0.18 & -0.76 & .445 \\
Formative Assessment        & -0.02 & 0.18 & -0.14 & .889 \\
Metacognition               & -0.06 & 0.18 & -0.35 & .724 \\
Power Dynamics              &  0.08 & 0.18 &  0.45 & .650 \\
\midrule
Group Var (participant)     &  0.11 & 0.05 & & \\
\bottomrule
\end{tabular}
\caption{Linear mixed-effects model coefficients. Reference categories: 
\textsc{Mistral-7B} (FT) for model, Accountability for dimension.}
\label{tab:lmm_results}
\end{table}

\paragraph{Post-hoc comparisons.} Tukey's HSD test confirms that 
\textsc{Gemini} and \textsc{Mistral} differ significantly 
(meandiff $= 0.331$, $p = .020$), while \textsc{Claude} vs.\ 
\textsc{Gemini} (meandiff $= 0.266$, $p = .078$) and \textsc{Claude} 
vs.\ \textsc{Mistral} (meandiff $= -0.065$, $p = .857$) do not. 
Pairwise Welch's $t$-tests with Bonferroni correction ($\alpha = 
.017$) confirm only the \textsc{Gemini} vs.\ \textsc{Mistral} 
difference ($t = 2.789$, $p = .006$, $d = 0.303$, small effect).

\end{document}